\documentclass[letterpaper]{article}
\usepackage[pass]{geometry}
\usepackage[implicit=false]{hyperref}
\usepackage{jair, theapa, rawfonts}
\usepackage{url}
\usepackage{graphicx}
\usepackage[section]{placeins}

\usepackage{amsmath,amsfonts,bm}

\def\eqref#1{equation~\ref{#1}}

\def\1{\bm{1}}

\DeclareMathAlphabet{\mathsfit}{\encodingdefault}{\sfdefault}{m}{sl}
\SetMathAlphabet{\mathsfit}{bold}{\encodingdefault}{\sfdefault}{bx}{n}

\jairheading{1}{1993c}{1-15}{6/91}{9/91}
\ShortHeadings{Set-to-Sequence Methods in Machine Learning}
{Jurewicz \& Derczynski}
\firstpageno{1}

\begin{document}

\title{Set-to-Sequence Methods in Machine Learning:\\a Review}

\author{\name Mateusz Jurewicz \email maju@itu.dk, mj@tjek.com\\
       \addr Department of Computer Science\\ IT University of Copenhagen, 2300
       Copenhagen, Denmark\\Tjek A/S, 1408 Copenhagen, Denmark
       \AND
       \name Leon Derczynski\email leod@itu.dk \\
       \addr Department of Computer Science\\ IT University of Copenhagen, 2300 Copenhagen, Denmark}


\maketitle

\begin{abstract}
Machine learning on sets towards sequential output is an important and \mbox{ubiquitous} task, with applications ranging from language modelling and meta-learning to multi-agent strategy games and power grid optimization. Combining \mbox{elements} of representation learning and structured prediction, its two primary challenges include obtaining a meaningful, permutation invariant set representation and subsequently utilizing this \mbox{representation} to output a complex target permutation. This paper provides a comprehensive introduction to the field as well as an overview of important machine learning methods tackling both of these key challenges, with a detailed qualitative comparison of selected model architectures.
\end{abstract}

\section{Introduction}

We begin by providing a definition of the set-to-sequence field and outline its importance in various areas of application.

\subsection{What is Set-to-Sequence?}

Set-to-sequence encompasses a group of problems where input takes the form of unordered collections of elements and the output is an ordered sequence. These challenges can be approached as a machine learning problem, where models learn arbitrary functions for performing the set-to-sequence mapping.

Set-to-sequence covers combinatorial optimization and structure prediction problems where exhaustive search is often not tractable. Machine learning (ML) approaches to set-to-sequence combine set-encoding techniques with permutation learning and have found an exceptionally wide range of practical applications.

Many of the successful deep learning approaches take advantage of the structure in their input data. However, sets do not posses the kind of internal structure that images and natural language sentences do. In set-to-sequence our input data does not have an inherent order and therefore our models must take into consideration the \textit{permutation invariance} of sets. Obtaining meaningful permutation invariant representations is an important challenge for machine learning models in order to ensure that the same set will not result in two different outputs, due to the arbitrary initial order in which its elements were presented to the model.

\subsection{Why Does Set-to-Sequence Matter?}

Machine learning set-to-sequence methods can approximate solutions to computationally intractable problems in many areas. They have been applied to learning competitive solvers for the NP-Hard Travelling Salesman Problem \shortcite{Vinyals2015}; tackling prominent NLP challenges such as sentence ordering \cite{Wang2019} and text summarization \shortcite{Sun2019}; and in multi-agent reinforcement learning \shortcite{Sunehag2018}. A notable example is the agent employed by the AlphaStar model, which defeated a grandmaster level player in the strategy game of Starcraft II \shortcite{Vinyals2019}. 

Set-to-sequence ML models also play an important role in data-intensive 3D point cloud processing \shortcite{Qi2017} and meta-learning \shortcite{Huang2018}. Set-input and set-ordering problems themselves are prominent in a wide array of applications ranging from power grid optimization \shortcite{Cui2019}, where solving them led to power usage savings of up to 30\%, through anomaly detection \shortcite{Jung2015} to measurements of contaminated galaxy clusters \shortcite{Ntampaka2016}.

This review contributes to the field in two primary ways:

\begin{enumerate}
    \item By providing a single point of entry for researchers interested in the set-to-sequence field and for applied practitioners solving set-input ordering challenges.
    \item By comparing all the discussed methods via a number of aspects relevant for both academic and applied work and presenting this comparison in the form of easy-to-read tables, which will help guide the reader towards the most applicable method for their specific area of interest.
\end{enumerate}

The remainder of this paper is structured in the following way: firstly, we introduce the reader to the necessary background concepts and related work in section \ref{background}, including specific notes on the adopted notation. Secondly, an overview of set encoding methods is given in section \ref{set-encoding-methods}, with comparison tables and details of the underlying mathematical transformations. Thirdly, section \ref{orderingmethods} contains a survey of popular ordering methods, which use the encoded set representation to output a complex permutation. The lists provided in sections \ref{set-encoding-methods} and \ref{orderingmethods} are not exhaustive and focus primarily on deep learning approaches. Finally, a discussion of current limitations and directions for further research is given in section \ref{discussion}, followed by a short conclusive summary in section \ref{conclusion}.

\section{Background} \label{background}

In this section the reader is introduced to the key concepts related to machine learning on sets and permutation learning, with minor notes on notation throughout the rest of the paper. Additionally, a comprehensive overview of related work from other fields of machine learning is given, including natural language processing, information retrieval and set segmentation.

\subsection{Important Concepts} \label{important_concepts}

For the purposes of this review a \textit{set} can be intuitively defined as a collection of distinct elements, without a canonical order between them \cite{halmos2017naive}. An important property of sets is that they can have a varying number of elements, also referred to as their \textit{cardinality}. Whilst the intuitive definition of a set is susceptible to known paradoxes \cite{rang1981zermelo}, the machine learning methods discussed here do not require an in-depth understanding of the proper definition from axiomatic set theory. Interested readers can find further information pertaining to it in other referred publications \cite{takeuti2013axiomatic}. 

As per the axiom of extensionality, sets are defined only by their elements \shortcite{hayden1968zermelo}. In practice this means that given, for example a set $A = \{x, y, z\}$ and set $B = \{ z, y, x\}$ we know that $A = B$. The order in which the elements are presented in roster notation does not matter. From now on, when we refer to a set, we specifically limit our considerations to finite sets only.

Set-to-sequence ML methods are distinctly different from earlier, encoder-decoder sequence-to-sequence model architectures \shortcite{Sutskever2014}. The difference stems from having to handle set-input data. This imposes two requirements on set-to-sequence ML methods that are not met by most neural network models:

\begin{enumerate}
    \item \textit{Permutation Invariance}
    
    The output of the model must be the same under every possible permutation of the elements from the input set.
    \itemsep0.5em
    
    \item \textit{Varying Input Length} 
    
    The same model must be able to process input sets of different lengths.
\end{enumerate}

If these criteria are not met, the ML model by definition treats its input as a sequence, not a set. Fully feed-forward methods fail to meet both criteria and the recurrent neural networks (RNNs), which form the foundation of most sequence-to-sequence autoencoders, are sensitive to alterations of the order of their input \shortcite{Vinyals2016}. To truly treat input data as an inherently unordered set we must be certain that permuting the input will not result in a different encoded set representation \shortcite{Zaheer2017}. Additionally, depending on the presence and type of a downstream task that uses this representation, we are interested in whether the final output is also permutation invariant, which is not necessarily the case with all reordering methods.

More formally, a function $f: \mathcal{P}(X) \rightarrow Y$ is permutation invariant regarding the order of the elements of its input set if for every permutation $\pi$ the following property holds: $f(\{x_1, \ldots , x_n\}) = f(\{x_{\pi(1)}, \ldots,  x_{\pi(n)}\})$. A related property of functions on sets, which has been formally investigated by \citeauthor{Zaheer2017} \citeyear{Zaheer2017}, is \textit{permutation equivariance}. In tasks where each set element has an associated target label, such that these individual labels depend on the entirety of the set, we would ideally want our predicted labels to remain the same per element, regardless of how the original input set is permuted. That property is permutation equivariance.

At this point it is important to distinguish between two types of set-to-sequence challenges. In the first type the output is a reordering of the input elements, with the possibility of repeating an element multiple times in the output sequence or skipping it entirely. Such permutations with potential repetition and exclusion are further denoted as \textit{complex permutations}. We can refer to the type of problems involving various kinds of permutations of the input elements as \textit{self-referential} set-to-sequence challenges. 

The self-referential set-to-sequence domain includes classic combinatorial optimization problems and forms the majority of this review. There are many different ways to frame this task and formalize the resulting output, which are discussed in section \ref{orderingmethods}. They include primarily pointer-based attention (\ref{attention-based-pointing}), the generation of permutation matrices (\ref{permutation-matrices}) and ranking scores (\ref{listwise-ranking}).

\begin{figure}[ht]
\begin{center}
  \includegraphics[scale=0.2]{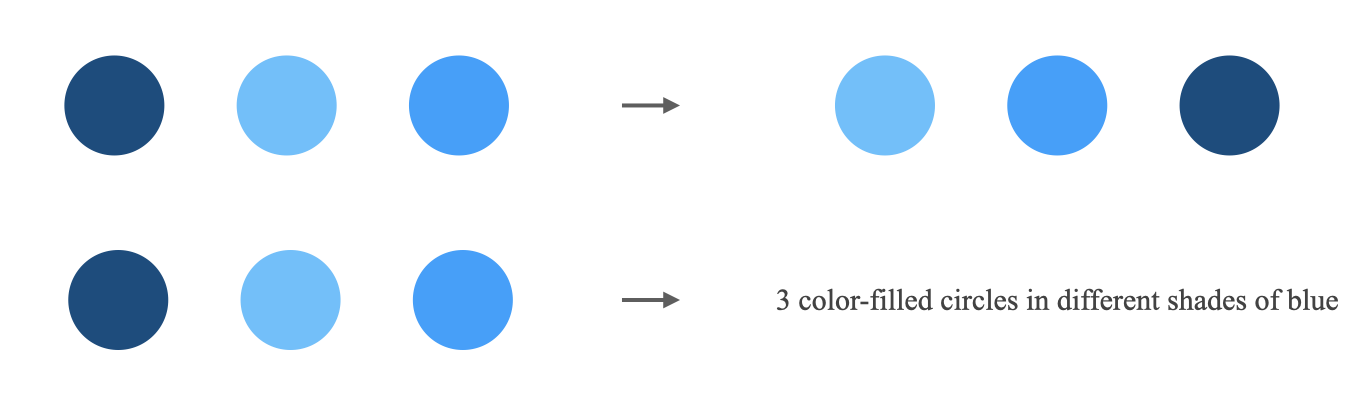}
  \caption{Set-to-Sequence Tasks by Referentiality}
  \label{illustration-1-types}
  \end{center}
\end{figure}

In the second type of set-to-sequence challenges, the task is to generate output that is sequential in nature, but is not defined as a permutation of the original input elements. We denote this as \textit{non-referential} set-to-sequence. It encompasses for example summarization of a set of documents. Here, the input is indeed a set, with unique elements that do not have a canonical ordering to them. The output is a sequence of natural language tokens in the form of a human-readable summary, without referring directly to the elements of the input set. This area is only partially covered by this review.

For a simple visual explanation highlighting the difference between self-referential (top) and non-referential (bottom) set-to-sequence tasks, see Figure \ref{illustration-1-types}. In both cases the input is a set of 3 disks of varying shades of blue. In a self-referential setting the target might be a permutation from lightest to darkest. In a non-referential setting the target may be a sequence of word tokens describing the input set. 

Specifically, all descriptions of set-input encoding methods are shared between the two types of set-to-sequence problems and therefore will be of value to readers interested in either. However, for considerations related to sequence prediction in general, areas of interest include recurrent neural network encoder-decoder models \cite{Sutskever2014}, reinforcement learning actor-critic methods \shortcite{Bahdanau2017RL} and fully-connected transformer architectures as investigated by \shortciteauthor{Vaswani2017} \citeyear{Vaswani2017},  \shortciteauthor{devlin-etal-2019-bert} \citeyear{devlin-etal-2019-bert} and \shortciteauthor{Brown2020LanguageMA} \citeyear{Brown2020LanguageMA}.

In summary, the former type of set-to-sequence ML architectures, which this review focuses on, tackle two primary challenges:

\begin{enumerate}
    \item Handling varying-length set-input data in a way that ensures permutation invariance
    \item Handling outputs as complex permutations or \textit{reorderings} of the original input
\end{enumerate}
    
The first challenge, once solved, allows us to use machine learning methods to perform set-input regression, classification, recommendation \cite{Vartak2017}, as well as clustering problems and more \cite{Edwards2016}. 

Depending on the specific task at hand, the permutation invariant representation of the input set may also be required to encode higher order interactions between the input set elements, as seen in the work of \shortciteauthor{Lee2019} \citeyear{Lee2019} and \shortciteauthor{Zhang2020} \citeyear{Zhang2020}, which is a separate but important consideration in the area of encoding sets. 

The second challenge of \textit{permutation learning} is made simpler by solving the first one, but has also been successfully tackled without addressing it \cite{Vinyals2015}. It focuses on learning the proper order of arbitrary input elements. As a result, the model learns to predict the best structure of the output sequence composition.

\subsection{Difficulty of Learning to Reorder}

Permutation learning is an inherently difficult challenge. Even a relatively simple application of set-to-sequence methods to the Travelling Salesman Problem (TSP) in a two dimensional Euclidean space involves tackling an NP-Hard problem \cite{Vinyals2015}.

Whilst highly successful, polynomial time algorithms for obtaining good approximate solutions to this task do exist, such as the ones proposed by \citeauthor{arora1996polynomial} \citeyear{arora1996polynomial} and \shortciteauthor{Karlin2020AI} \citeyear{Karlin2020AI}, it is also valuable to investigate the capacity of current machine learning techniques to learn them iteratively.

\begin{figure}[ht]
\begin{center}
  \includegraphics[scale=0.2]{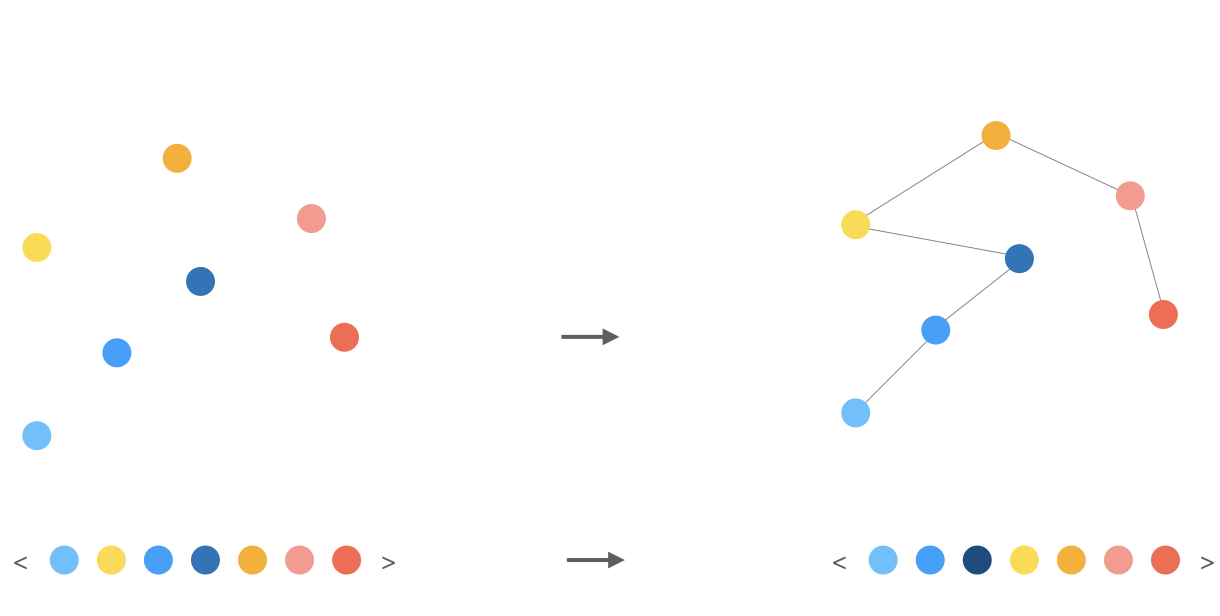}
  \caption{Travelling Salesman Problem}
  \label{illustration-2-tsp}
  \end{center}
  \small An example of a TSP input is given as a set of points in a 2D Euclidean space (top left) and an arbitrarily ordered array (bottom left). Individual dots represent $x, y$ coordinates. As output, we see the shortest path between points (top right) and an array representing their target order (bottom right).
\end{figure}

In the two dimensional Euclidean version of the TSP our input is a set of point coordinates and our desired output is a permutation of these points in a way that results in the shortest distance travelled between them. Additionally, we must not skip nor revisit any of the points. For a visual example, see Figure \ref{illustration-2-tsp}.

This challenge is difficult because the number of possible permutations increases factorially in the cardinality of the input set. It is a self-referential set-to-sequence problem due to its input having no inherent order and the desired output being a permutation of the input elements. Given the same set of points as input, we expect the output to be the same tour, regardless of the order in which they are originally presented. 

Further difficulties arise when we consider how to formalize the resulting reordering. One possible method involves the use of the aforementioned permutation matrices, which are discrete and therefore do not lend themselves to direct use of gradient-based backpropagation without a relaxation of the concept \cite{Emami2018LearningPW}. This and other formulations of representing a reordering (mentioned below) are discussed in more detail in section \ref{orderingmethods}.

Such difficulties have motivated researchers to instead investigate differentiable, attention-based methods that involve pointing to the elements of the original set to define their permuted order. This approach often involves potentially computationally expensive beam search during inference.

Finally, if we forego the requirement of handling varying-length inputs, we can look to traditional learn-to-rank approaches for inspiration. In such frameworks the reordering is formulated as the assignment of a relevance score to each element, followed by sorting the elements according to that score, in monotonic order. However, sorting is a piecewise linear function, which therefore may contain many kinks where it is not differentiable. As a result, differentiable proxies to the sorting operator have been developed, but they did not achieve the expected $O(n~\textrm{log}~n)$ time complexity until a method consisting of a projection onto a \textit{permutahedron} was proposed by \shortciteauthor{Blondel2020FastDS} \citeyear{Blondel2020FastDS}.

Alternatively, in learn-to-rank, our model may be trained to return \textit{ranks}, i.e. positions of the input elements in the target (properly ordered) sequence. These ranks are piecewise constant functions, with derivatives that are either null or undefined, preventing gradient-based learning. However, significant progress has been made towards directly approximating ranking metrics \shortcite{rolinek2020optimizing} and constructing differentiable sorting and ranking operators \cite{Blondel2020FastDS}. Additionally, \shortciteauthor{engilberge2019sodeep} \citeyear{engilberge2019sodeep} propose a deep neural net which can act as a differentiable proxy for ranking, allowing the use of traditionally non-differentiable metrics such as Spearman's rank-order correlation \cite{spearman1904proof} as loss functions.

\subsection{ML on Sets and Combinatorial Optimization}

Set-to-sequence combines techniques from the field of machine learning on sets and combinatorial optimization. The former covers research areas related to both set-input and set-output problems, of which set-to-sequence is only concerned with the first kind. The latter consists of finding an optimal object from a finite set of objects and is strongly related to many forms of ordering tasks. The canonical example is the aforementioned TSP, which in itself has a long history of attempts at solving it through the most popular machine learning methods of the time, for example \citeauthor{Smith1999} \citeyear{Smith1999}, \citeauthor{pihera2014application} \citeyear{pihera2014application}, \shortciteauthor{ishaya2019comparative} \citeyear{ishaya2019comparative} and \shortciteauthor{Bengio2020} \citeyear{Bengio2020}.

Combinatorial optimization as such is of vital importance to modern industry applications. Consider the archetypal \textit{Vehicle Routing Problem} (VRP), which poses the task of finding an optimal set of routes for a fleet of vehicles aiming to deliver goods to a given set of locations. The quality of the solution is determined by the \textit{global transportation cost}. In the simplest variant of VRP, this is dependent on the sum of the lengths of tours for all vehicles. This effectively requires an ordering of the locations into optimal trips, per each vehicle. Given the scale of modern logistical challenges and the environmental impact of freight, it is understandable that there have been many attempts to apply recent machine learning developments to such problems \cite{Ibrahim2019}. 

Current state-of-the-art combinatorial optimization algorithms often rely on handcrafted and hard-to-maintain heuristics for making decisions that are otherwise computationally infeasible or not well defined mathematically, for example \shortciteauthor{bello2016neural} \citeyear{bello2016neural}. It is a natural area of application for machine learning research and has been approached through the use of graph-based methods \shortcite{Dai2017}, reinforcement learning \shortcite{Nazari2018} and attention mechanisms \shortcite{Kool2019}. For a comprehensive survey of the wider intersection of combinatorial optimization and machine learning, see \citeauthor{Bengio2020} \citeyear{Bengio2020}.

\subsection{Notation}

The paper follows the notational conventions that are most common in literature. Scalar values are marked with lower case italics $x_i$, vectors with lower case bold typeface $\mathbf{x}$, matrices with capital case italics $X$. These matrices may be used to represent sets, in which case they are presented through roster notation with curly brackets, for example $X = \{ \mathbf{x_1}, \ldots, \mathbf{x_n}\}$.

However, sets may also consist of scalar elements, in which case a capital letter is still used to represent them: $A = \{a_i, \ldots, a_n\}$. Given the importance of differentiating between unordered sets and ordered sequences, the latter are represented through angled brackets $\mathbf{x} = \langle x_1, \ldots, x_n \rangle$ for additional clarity. When indicating the index of an element within a vector, whose symbol already contains a subscript (e.g. $\mathbf{v_j}$) the index of the scalar element is given in the superscript ($v^i_j$).

Individual permutations are marked as $\pi$, such that an example $\pi_i = \langle 3, 2, 1 \rangle$, consisting of integer indices referring to the original sequence $\mathbf{x} = \langle x_1, x_2, x_3\rangle$, would result in the reordered sequence $\mathbf{x^\pi} = \langle x_3, x_2, x_1 \rangle$. In some cases, if the order of elements in the original sequence is nontrivial, a permutation $\pi$ can also be given in two-line notation making both $\mathbf{x^\pi}$ and the integer indices explicit:

\begin{equation}
\pi = \begin{pmatrix}
x_3 & x_2 & x_1\\
3 & 2 & 1
\end{pmatrix}
\end{equation}

\subsection{Connections with Other ML Fields}

In this section, a brief overview of other related topics from different fields of machine learning research is given. The aim is to point the reader who may only be tangentially concerned with set-to-sequence tasks to the appropriate area within their main field of interest. A reader with a decided focus on set-to-sequence is encouraged to continue reading section \ref{set-encoding-methods} directly.

\subsubsection{Natural Language Processing (NLP) }

There is a number of cases from the field of \textit{Natural Language Processing} (NLP) that require tackling similar challenges to the ones faced by set-to-sequence methods. The popular sequence-to-sequence, encoder-decoder framework proposed by \citeauthor{Sutskever2014} \citeyear{Sutskever2014} can itself, in principle, be applied to set-to-sequence problems, but does not perform well in practice. An example of such a case is the work on word ordering tasks, also known as \textit{linearizations}, towards syntactically plausible word representations \cite{Nishida2017}. 

The authors use sentences of ordered words to train the network to output a binary permutation matrix. When the original input, in the form of randomly ordered words from the target sentence, gets matrix-multiplied by this permutation matrix, the proper order is recovered. The network, referred to by the authors as the \textit{Word Ordering Network} (WON), is an example of one way to formalize an ordering task. It can be seen as a simplification of the \textit{Pointer Network} encoding method \cite{Vinyals2015}, discussed in more depth in a later section, whilst more closely resembling the classic sequence-to-sequence models in the decoder (where it sequentially outputs rows of the permutation matrix).

Another example of an NLP task where permuting plays a key role is sentence ordering and order discrimination. The goal is to take a set of sentences and order them back into the original paragraph. Historically, this area of research has been dominated by hierarchical RNN-based approaches, which make use of LSTMs or GRUs in an auto-encoder framework \shortcite{Logeswaran2018}. First, a recurrent network is used to obtain the embedding of each sentence and then another to obtain the context representation of the entire paragraph. 

More recent developments have seen the use of attention mechanisms to make it easier for this embedding to encode vital information regardless of the distance between information-carrying elements, as per the vanishing gradient problem. An example of this can be found in the \textit{ATTOrderNet} architecture \shortcite{Cui2018} and more recently in the \textit{Set Transformer} \cite{Lee2019}, discussed in detail in section \ref{set-transformer}.

An example of an NLP set-to-sequence challenge where the output is not a permutation is the task of summarizing multiple documents into a single sequence of text as seen in the works of \shortciteauthor{ma2016unsupervised} \citeyear{ma2016unsupervised} and \shortciteauthor{mani2018multi} \citeyear{mani2018multi}.

\subsubsection{Ranking, Information Retrieval and Content Ordering}

Another related, but succinctly distinct field comes in the form of ranking problems, information retrieval and content ordering. These encompass a family of challenges where there exists an \textit{optimal} hierarchical order to the input elements, such that given two elements there is always a proper way in which they should be placed in relation to one another, which does not change depending on the other input-set elements.

More specifically, in the context of ranking for search, if we enter the query `cats`, the returned image of a cat should always rank higher than an image of a `dog`, regardless of what the other returned images may contain. This assumption is not always true in complex set-to-sequence problems, where any new element of the input-set can change what the proper relative order or structure of the already available elements' sequence should be.

Traditionally, learn-to-rank problems have been tackled in a pairwise manner \shortcite{Cohen1998}, later approaches have applied neural methods on a list-based formulation of this problem \shortcite{cao2007learning}. Ranking has also found application in content selection \shortcite{Puduppully2019} and been employed as a useful auxiliary objective in a multitask setting for regression problems \shortcite{Liu2019ranking}. A detailed look at listwise ranking approaches to ordering and structure prediction can be found in section \ref{listwise-ranking}.

\subsubsection{Set Regression, Classification and Segmentation}

A more closely related area of work stems from set-input problems that have an output that is not a sequence. These include set regression, classification and segmentation challenges, among others. Effectively, this research field requires solving a near identical challenge to the first of two primary set-to-sequence challenges outlined at the beginning of this section, in that obtaining a proper encoding of the input set is vital. 

Examples of such methods include \textit{PointNets} for 3-dimensional point cloud classification and segmentation \cite{Qi2017}, which builds on previous work by \citeauthor{Vinyals2016} \citeyear{Vinyals2016} in a specific geometric setting requiring both rotation and translation invariance (see section \ref{point-net}). Another example comes in the form of techniques for labelling objects based on a set of images from multiple viewpoints such as security cameras \shortcite{Zhao2019ARO} and even fully convolutional models for set segmentation \shortcite{Oliveira2020}. 

Methods that obtain the input set representation in a way that is interesting to set-to-sequence problems are included in the main section of this review and given appropriate focus, regardless of whether their original application was in sequence-output challenges.

\subsubsection{Set-Output Tasks, Including Set-to-Set}

Conversely, a large field of work revolves around methods that learn to generate or predict a set as their output. Sets are the natural representation for many kinds of output data in machine learning tasks. These include a set of objects present in an image in an object detection setting \shortcite{Zhao2019detection}, a group of points in a point cloud \shortcite{Achlioptas2018} and a selection of nodes in a molecular graph for the problem of molecule generation \cite{DeCao2018}. 

The main challenge of set-output methods mirrors the primary challenge of representing sets in a permutation invariant way in set-input problems. If the order in which the model outputs elements does not matter, there are $n!$ equivalent, correct outputs that the model has to learn to consider equally good. 

For example, imagine a simple task where the model must learn to take as input a set of integers and return the set of all primes present in the input. Specifically, given the input $A = \{1, 2, 3, 4, 5\}$ the correct output should take the form of the set $B = \{2, 3, 5\}$. However, given that ML model implementations operate on ordered, multidimensional arrays, the model must learn to treat all of these possible output sequences as equally correct: $\langle 2, 3, 5 \rangle, \langle 2, 5, 3\rangle, \langle3, 2, 5\rangle, \langle 3, 5, 2\rangle, \langle 5, 2, 3\rangle, \langle 5, 3, 2\rangle$.

Failure to properly account for this property of output sets leads to discontinuities that are difficult for most modern neural architectures to learn, even on seemingly trivial synthetic datasets \cite{Zhang2020}. An example that illustrates this comes in the form of an autoencoder trained to embed and then reconstruct input consisting of a set of $n$ 2-dimensional points forming a regular polygon. 

Every example in this dataset is a rotation of the same polygon around the origin. The discontinuity arises from this rotation, which forces a switch with regards to which element of the input set the model's output neurons will be responsible for decoding. This is referred to as the \textit{responsibility problem}. Proper handling of the set structure in the output requires the application of permutation invariant and permutation equivariant operations, much like in set-input problems, where the responsibility problem is not present.

Notable recent methods in the field of set prediction include the \textit{Deep Set Prediction Network} (DSPN) by \shortciteauthor{Zhang2019dspn} \citeyear{Zhang2019dspn}, which consists of a deep learning vector-to-set model that enables permutation invariance, and the \textit{Transformer Set Prediction Network} (TSPN) by \shortciteauthor{Kosiorek2020} \citeyear{Kosiorek2020}, that additionally takes advantage of the multiheaded self-attention introduced by \citeauthor{Vaswani2017} \citeyear{Vaswani2017}. The TSPN addresses the limitations of the DSPN related to set-cardinality learning. Additionally, an iterative attention mechanism referred to as \textit{Slot Attention} has been proposed by \shortciteauthor{Locatello2020} \citeyear{Locatello2020}, which decomposes input features into a set of representations, lending itself to set prediction tasks.

A sub-field of interest within set prediction is referred to as set-to-set, where both the input and output are structured as a set. Such tasks include recommendation \shortcite{Sarwar2001}, image search \shortcite{Wang2014} and person re-identification \shortcite{Zheng2015}. Set-to-set challenges require both permutation invariance in the parts of the model that encode the input set and a proper cross-similarity function for the output sets \shortcite{Saito2019DeepSM}, circumnavigating the responsibility problem.

\subsubsection{Ensuring Other Types of Invariance}

Finally, it may be of value to mention methods that obtain types of invariance other than under permutation. These methods stem from areas of application where the model's final prediction should not be dependent on such predefined transformations. Examples include translational and rotational (also known as viewpoint) invariance, common in computer vision problems, addressed in the works of \shortciteauthor{ling2016machine} \citeyear{ling2016machine} and \citeauthor{marcos2016learning} \citeyear{marcos2016learning}. Taking as illustrative case the task of object detection, to ensure the former quality the model needs to recognize the same object regardless of its position within the input image. To ensure the latter, given three dimensional images of a scene, the model must correctly identify an object regardless of the angle from which it is being perceived.

An important traditional approach to learning models that are invariant to certain transformation is data augmentation, as seen in \citeauthor{Taylor2018ImprovingDL} \citeyear{Taylor2018ImprovingDL} and \shortciteauthor{HernndezGarca2019LearningRV} \citeyear{HernndezGarca2019LearningRV}. Here, the underlying idea is that the model will be able to learn to become invariant to the chosen transformations once we augment our data with instructive examples that have been appropriately translated, rotated, partially obscured, blurred, illuminated, resized or had other applicable transformation applied to them, often in combination. More recent approaches that aim to safeguard machine learning models against relying on unintended aspects of data through adversarial strategies have also been proposed \shortcite{jaiswal2018unsupervised}. In order to apply the data augmentation approach to the set-to-sequence domain, we would increase our training set up to $n!$ times, providing the model with every possible permutation of each example set. 

To prevent the costs associated with a larger training set, machine learning methods commonly employ various pooling operators after the stacked, equivariant feature extraction layers to obtain the desired invariance. However, in the case of translation invariance through pooled convolutional operations, the assumption that this completely prevents the model from exploiting the absolute location of an object in an image has been challenged \cite{kayhan2020translation}.

\section{Set Encoding Methods} \label{set-encoding-methods}

In this section, a qualitative comparison of different set encoding ML methods is provided, followed by detailed subsections devoted to the individual model architectures.

\subsection{Method Comparison} \label{method-comparison-set-encoding}

This section introduces the reader to each of the relevant set-encoding methods in turn. These are also sometimes referred to in literature as \textit{set-pooling} methods \cite{Lee2019}. Additionally, a number of comparison tables provides a summary overview: \ref{set-encoding-table-1}, \ref{set-encoding-table-2}.

It is important to note that some of the methods discussed in this section were designed specifically to handle set-to-sequence problems. As such, they contain both a set encoding module and a permutation outputting mechanism. Since it is not always immediately obvious how to combine a method that handles set encoding with a method that is designed to output a reordering, we compare various aspects of these methods in multiple places. 

The models are compared with regards to the following aspects:

\begin{enumerate}

    \item \textbf{Permutation Invariance}: whether the model obtains a permutation invariant representation of the input set. The same set must result in the same embedded representation, regardless of how the actual input array was permuted.
    
\end{enumerate}

This feature does not guarantee that the final output of the model will be the same for differently ordered sequences obtained from a single set, as that may depend on the ordering method applied to the obtained permutation invariant set embedding, in order to output a sequence. This stems from the fact that these methods may require the reintroduction of the information relating to the order of the original input array and refer to it directly when outputting a permutation - particularly pointer-based attention. For more information on this, see section \ref{orderingmethods}.

\begin{enumerate}
\setcounter{enumi}{1}

    \item \textbf{Multiset Input}: whether the model can distinguish between a given input set and certain examples of its corresponding multisets, with repeated elements. 

\end{enumerate}

For example, the  $\textrm{average}()$ pooling operator will not be able to distinguish between a set $X = \{1, 2, 3 \}$ and a multiset $X^\prime = \{1, 1, 2, 2, 3, 3\}$. The $\textrm{max}()$ operator will similarly fail in the case of $X^{\prime \prime} = \{1, 1, 2, 3\}$.

\begin{enumerate}
\setcounter{enumi}{2}
    
    \item \textbf{Complexity}: how the model's computational complexity relates to the cardinality $n$ of the input set, and possibly other hyperparameters specific to a given architecture.
    
    \item \textbf{Applications}: selected prominent domains which the model has been successfully applied to. Further examples can be found in the later sections devoted to each model. 

\end{enumerate}

Additionally, the most prominent architectures can be split into RNN-based methods, namely the Pointer Network and Read-Process-and-Write model, and the more recent fully connected ones, primarily the foundational DeepSets method and the Set Transformer. 

Alternatively, an interesting way to distinguish between them would be to consider methods that depend on a variation of the attention mechanism introduced by \shortciteauthor{bahdanau2014neural} \citeyear{bahdanau2014neural}, such as the Pointer Network and the Set Transformer, and others that do not utilize it.

\begin{table}[htb]
\vspace*{+5mm} 
\label{set-encoding-table-1}
\begin{center}
\begin{tabular}{lccc}
\multicolumn{1}{c}{\bf Model}  
&\multicolumn{1}{c}{\bf Perm. Invariance}
&\multicolumn{1}{c}{\bf Multiset}
&\multicolumn{1}{c}{\bf Complexity}
\\ \hline \\
Pointer Network \citeyear{Vinyals2015}&No &No\footnotemark[1] &$O(n^2)$ \\
Read-Process-and-Write \citeyear{Vinyals2016}&Yes &Yes &$O(n^2)$\\
PointNet \citeyear{Qi2017}&Yes &No\footnotemark[2] &$O(n)$\\
DeepSets \citeyear{Zaheer2017}&Yes &Yes\footnotemark[3] &$O(n)$\\
Janossy Pooling \citeyear{Murphy2019}&Yes &Yes &$O(n!)$\\
Set Transformer \citeyear{Lee2019}&Yes &Yes &$O(n^2)$\\
AttSets \citeyear{yang2020robust}&Yes &Yes &$O(n)$\\
FSPooling \citeyear{Zhang2020}&Yes &Yes &$O(n~\textrm{log}^2 n)$\\
RepSet \citeyear{Skianis2020}&Yes &Yes &$O(mn + n^2~ \textrm{log}~n)$\\
\end{tabular}
\end{center}
\caption{Comparison of set encoding methods, part 1}
\end{table}

\subsubsection{Notes on Complexity}
\label{notes-on-complexity}

The complexity of the Read-Process-and-Write method is additionally impacted by the number $t$ of steps in the \textit{Process} block that computes the permutation invariant embedding of the input set. $t$ is constant, but an interesting area of further research would be to learn it adaptively, similar to the method described by \citeauthor{Graves2016-adaptive-time} \citeyear{Graves2016-adaptive-time}. The complexity of PointNet becomes $O(n^2)$ for 2D images and $O(n^3)$ for voxels, due to the convolutional operations. The authors of Janossy Pooling propose 3 methods of balancing tractability and the model's representational power, as outlined in section: \ref{janossy-pooling}

Regarding the Set Transformer, the use of $l$ stacked SAB layers results in quadratic complexity of $O(n^2l)$, use of a stack of $l$ proposed ISAB layers with $m$ inducing points results in complexity of $O(nlm)$. Similarly to the PointNet architecture, the complexity of AttSets grows depending on the dimensionality of the input, due to the use of convolutional layers in the encoder. However, the authors also provide a novel training paradigm, called \textit{FASet}, and benchmark its mean time consumption for a single object against a selection of simple pooling methods, with favourable performance.

Regarding the RepSet method, $m$ is the chosen number of hidden sets for the bipartite matching algorithm, represented by the columns of a trainable matrix. For more information on this and a proposed, more tractable relaxation, see section \ref{rep-set}.

\renewcommand{\thefootnote}{\roman{footnote}}
\begin{table}[htb]
\vspace*{+5mm} 
\label{set-encoding-table-2}
\begin{center}
\begin{tabular}{llll}
\multicolumn{1}{c}{\bf Model}  
&\multicolumn{1}{c}{\bf Applications}
\\ \hline \\
Pointer Network \citeyear{Vinyals2015}&combinatorial, multi-agent \\
Read-Process-and-Write \citeyear{Vinyals2016}&combinatorial, sorting\\
PointNet \citeyear{Qi2017}& 3D shape classification and segmentation \\
DeepSets \citeyear{Zaheer2017}&set expansion, anomaly detection\\
Janossy Pooling \citeyear{Murphy2019}& arithmetic, graph classification\\
Set Transformer \citeyear{Lee2019}&amortized clustering, anomaly detection\\
AttSets \citeyear{yang2020robust}&3D shape reconstruction\\
FSPooling \citeyear{Zhang2020}&set and graph classification\\
RepSet \citeyear{Skianis2020}&text and graph classification\\
\end{tabular}
\end{center}
\caption{Comparison of set encoding methods, part 2}
\end{table}

\renewcommand\thefootnote{\arabic{footnote}}
\footnotetext[1]{The Pointer Network does not treat its input properly as a set, therefore it cannot be said to properly handle multiset input either, but it will distinguish between input vectors with repeated elements.}
\footnotetext[2]{Due to experimental results on the selected tasks, the authors of PointNet settle on max() as their recommended pooling operator, which does not distinguish between certain multiset variants. However, they report robust measurements of the performance of other pooling methods which can easily be included in the final model architecture and provide comparable results.}
\footnotetext[3]{Depends on the pooling operator used after the stacked fully-connected layers, of which the authors of DeepSets primarily focus on sum(), which does distinguish between sets and multisets. However, max(), which does not is also proposed as a problem-dependent variation. The formal proof extending DeepSets to multiset inputs was given by \shortciteauthor{Xu2019} \citeyear{Xu2019}.}
\renewcommand{\thefootnote}{\roman{footnote}}

\subsubsection{Notes on Dataset Performance}\label{notes-on-performance}

The listed set encoding methods can be applied to a wide spectrum of tasks. As a result, their performance has been tested on a variety of datasets, often in subtly different settings, which prevents direct comparison. In lieu of a table presenting their performance on a selected subset of such datasets, we provide a short discussion of the experimental results that do lend themselves to being compared. A more comprehensive experimental analysis in this area is a possible direction for future work.

Both Pointer Networks and the Read-Process-and-Write (RPW) method have been tested on the simple task of sorting a set of five floating point numbers between 0 and 1 \cite{Vinyals2016}. The Pointer Network achieved an accuracy of 0.90 compared to 0.94 reached by the RPW. Additionally, the RPW method appeared to be better at handling larger sets of floats. Both DeepSets and Janossy Pooling \cite{Murphy2019} have been tested on simple arithmetic tasks such as sum-of-digits prediction, with each method reaching an accuracy of 1.0, albeit tested on input sets of different cardinalities. The Set Transformer has instead been tested on maximum value regression.

The Set Transformer, DeepSets and Janossy Pooling have also all been tested in terms of performance on unique count tasks. However, in the case of the Set Transformer experiments were performed on sets of handwritten characters from the Omniglot dataset \shortcite{lake2019omniglot}, in the case of DeepSets on the MNIST8m hand-written digits \shortcite{loosli2007training} and in the case of Janossy Pooling on simple integer sets.

The most commonly shared experimental task in the papers introducing and consequently utilizing the listed methods was point cloud classification. Particularly the ModelNet40 dataset \shortcite{wu20153d} has been used to test four of the mentioned models. Whilst AttSets \cite{yang2020robust} employs it to formulate a multi-view reconstruction task, the other three methods are tested on the core classification task with PointNet reaching an accuracy of 0.892 \cite{Qi2017}, DeepSets  0.900 \cite{Zaheer2017} and the Set Transformer 0.904 \cite{Lee2019}. However, the specific methods used to produce the point clouds from the provided mesh representation of objects showcased certain differences, further highlighting the need for a systematic, uniform comparison.

Both the Set Transformer and DeepSets methods have been tested on the task of set anomaly detection, specifically by way of the CelebA dataset \shortcite{liu2015faceattributes}. However, the DeepSets model was tested in terms of accuracy (0.75) and the Set Transformer in terms of the area under receiver operating characteristic curve and area under precision-recall curve, preventing direct comparison.

The FSPool technique's performance has been compared to the Janossy Pooling method through a visual question answering task, employing the CLEVR \shortcite{johnson2017clevr} dataset. The accuracy of the latter was reported as 0.97 $\pm$ 0.54, and of the former as 0.9927 $\pm$ 0.18 \cite{Zhang2020}.

Another useful task for the purposes of performance comparison is document classification, where given a document, the input to the model is the set of embeddings of its terms. DeepSets, Set Transformer and RepSet have been directly compared through their performance in this regard on 8 separate datasets \cite{Skianis2020}, with the Set Transformer consistently outperforming DeepSets, and RepSet outperforming both of the aforementioned methods. 

Finally, the performance of DeepSets, Set Transformer and RepSet has been compared on the task of graph classification through the 5 datasets proposed by \shortciteauthor{kersting2016benchmark} \citeyear{kersting2016benchmark}. The classification accuracy of DeepSets on the MUTAG dataset was 0.862, Set Transformer's was 0.877 and RepSet's 0.886. However, on the arguably more difficult IMDB MULTI dataset the Set Transformer outperformed RepSet, reaching an accuracy of 0.502, compared to 0.499. For a full overview, see the paper by \citeauthor{Skianis2020} \citeyear{Skianis2020}. 

Further details regarding the performance and limitations of presented methods are available in the sections devoted to them individually (below).

\subsection{Pointer Networks} \label{pointer-networks}

The Pointer Network \cite{Vinyals2015} is an encoder-decoder neural network architecture including a modified attention mechanism, which allows it to learn a target reordering of input elements. It is the first deep learning method capable of taking sets as input and learning a desired permutation, resulting in complex output sequences.

Pointer Networks were originally designed to tackle combinatorial optimization problems with varying input sizes, which was their main advantage over previous sequence-to-sequence methods. A Pointer Network can be trained on inputs of varying length and has been demonstrated to generalize reasonably well to unseen lengths \cite{Vinyals2015}. 

Additionally, Pointer Networks included a modification of the content-based attention mechanism introduced by \citeauthor{bahdanau2014neural} \citeyear{bahdanau2014neural} which made it possible to treat the output of the model as pointers to elements of the input sequence. This attention-based pointing is one of the most popular methods for giving models the ability to output a permutation of the original input, regardless of the way they encode the original set. Due to its importance as a purely element-ordering technique, it is separately described in further detail in section \ref{attention-based-pointing}.

\subsubsection{Pointer Networks Limitations} \label{pointer-networks-limitations}

An important characteristic of Pointer Networks is that they do not strictly treat the input as a set, instead processing it solely through sequential recurrent neural networks. As a partial consequence they do not obtain a permutation invariant representation of the encoded set. This results in a situation where the same set can be represented as two differently ordered input arrays, leading to the model predicting two different outputs for it. Thus returning the optimal order is not guaranteed. 

Another important limitation is that nothing is explicitly preventing the model from outputting an invalid reordering of the input set or sequence. This becomes apparent during early training, when the model points to the same elements of the input at various indices of the output sequence. However, this can be mitigated by the addition of a beam search mechanism to the decoder during inference or by progressive masking. In the latter case, the entry in the attention vector referring to an element that had already been pointed to is preset to an infinitely negative value at each successive iteration, preventing it from being pointed to again, at the cost of certain inductive bias being introduced into the model.

\subsubsection{Pointer Networks Details} \label{pointer-networks-details}

The Pointer Network consists of a recurrent neural network (RNN) encoder and an RNN decoder with a modified attention mechanism. The model obtains a content-based attention vector $\mathbf{a_j} \in \mathbb{R}^n$ at each decoder step $j$. This vector represents the conditional probability of each input element $x_i$ being the correct one to be pointed to at this step, conditioned on all previous steps as well as the entire input sequence $\mathbf{x} = \langle x_i, \ldots, x_n \rangle$, in the form of all encoder hidden states $E = \langle \mathbf{e_1}, \ldots, \mathbf{e_n} \rangle$ obtained when the encoder block iterates over the input array. 

For simplicity, we will assume that each element $x_i$ must be pointed to exactly once, meaning that the output sequence of nonnegative integer pointers $\mathbf{y} = \langle y_i, \ldots, y_n \rangle \in \mathbb{Z}^n$ represents a valid permutation $\pi$, such that a sample target output $\mathbf{y}^\pi = \langle 0, 2, 1 \rangle$ would represent the reordered sequence  $\mathbf{x}^\pi = \langle x_1, x_2, x_3 \rangle$ for the sample input $\mathbf{x} = \langle x_1, x_3, x_2 \rangle$. This will mean that when iterating over both encoder states $\mathbf{e_i}$ and decoder states $\mathbf{d_j}$, we will always be in range $1$ to $n$. The input sequence $\mathbf{x}$ can itself consist of multidimensional elements, or such embeddings of each $x_i$ can be obtained prior to the pointer network module through a chosen embedding layer.

The attention mechanism in the decoder block is as follows:

\begin{equation}\label{1}
z^i_{j} = \mathbf{v^T} \textrm{tanh}(W_1 \mathbf{e_i} + W_2 \mathbf{d_j})\qquad \quad \textrm{for}~i \in (1, \ldots ,n)
\end{equation}

\begin{equation}\label{2}
\ \mathbf{a_j} =  \textrm{softmax}(\mathbf{z_j})\qquad \qquad \quad \ \ \ \ \ \ \ \ \textrm{for}~j \in (1, \ldots ,n)
\end{equation}

\begin{equation}\label{3} \ P(y_i | y_1, \ldots, y_{i-1}, \mathbf{x}) = \mathbf{a_j} \ \ \ \ \ \ \ \ \ \ \ \ \   \textrm{for}~j \in (1, \ldots ,n)
\end{equation}

\begin{figure}[ht]
\begin{center}
  \includegraphics[scale=0.3]{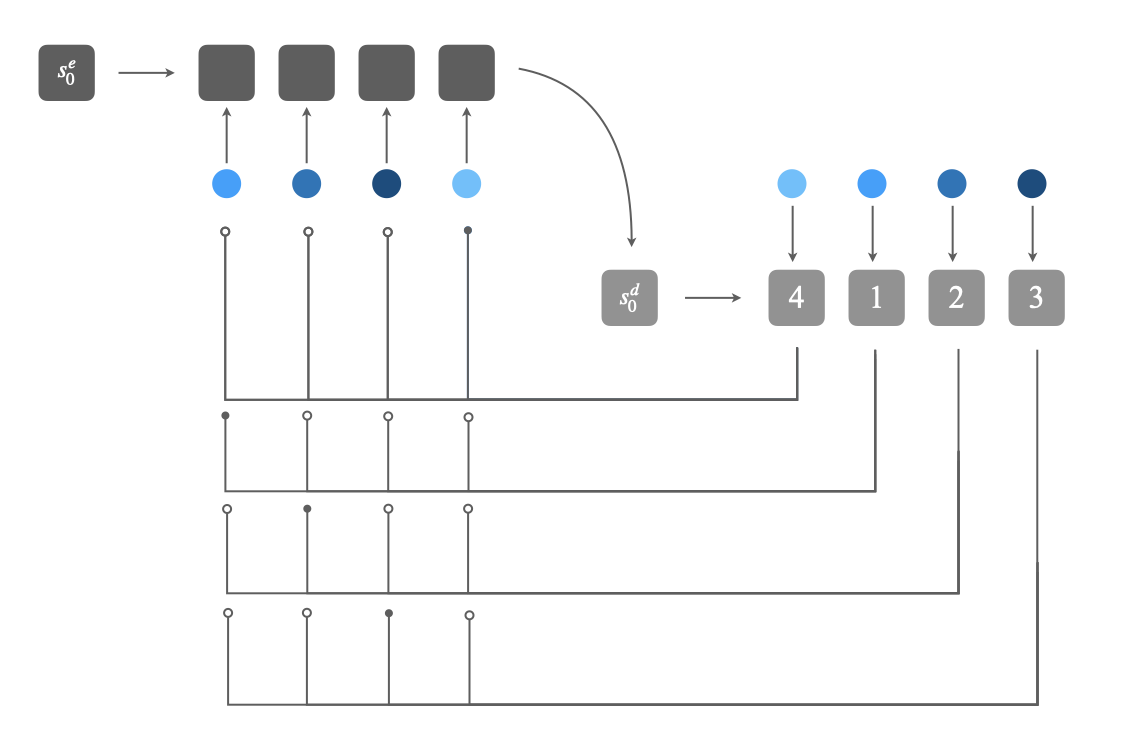}
  \caption{Pointer Network}
  \label{illustration-ptr-net}
  \end{center} 
  \small In Figure \ref{illustration-ptr-net}, an encoding RNN sequentially processes each element of the input array (blue dots), encoding it into a hidden state $s^e_n$ (dark gray), which is fed to the pointing decoder RNN (light gray). At every step, the second network produces a vector that represents a content-based pointer attention over the encoded inputs.
\end{figure}

Where $\mathbf{d_j}$ is the decoder's hidden state at the $j$-th output element, $\mathbf{e_i}$ is the encoder hidden state at the $i$-th input element, $W_1$, $W_2$ and $\mathbf{v}$ are trainable tensors.  The $\mathbf{z_j}$ vector is of the same length as the input $\mathbf{x}$ and represents an output distribution over the dictionary of input elements. After the application of the softmax nonlinear activation function, turning it into $\mathbf{a_j}$, it becomes an attention vector. For a visual explanation, see Figure \ref{illustration-ptr-net}.

\subsubsection{Pointer Networks Applications}

The primary application of Pointer Networks are tasks where the target output is a reordering of the elements of the initial input. This reordering is based on pointers to indices of the original input sequence. Examples of such problems in currently active research areas include element sorting, coherence modeling \cite{Logeswaran2018}, word ordering \cite{Cui2018} and sentence ordering \cite{Wang2019}, as well as summarization \cite{Sun2019} and ranking in information extraction \shortcite{Bello2018}.

In the original paper, the Pointer Network models have been tested on challenging combinatorial optimization problems such as finding planar convex hulls, computing Delaunay triangulations and the Travelling Salesman Problem. Experiments have shown that even with computationally intractable, NP-Hard problems such as TSP, this model architecture was able to learn competitive approximate solutions, limited by the scale of the problem, with $n <= 50$ for the TSP.

Pointer Networks also found usage within the AlphaStar reinforcement learning model which defeated a grandmaster level player in the competitive real-time strategy game of Starcraft II \cite{Vinyals2019}. They were employed to help the agent manage the structured, combinatorial action space in conjunction with an auto-regressive policy.

\subsection{Read-Process-and-Write Model}

The Read-Process-and-Write (RPW) model is a neural network architecture consisting of three distinct blocks and aiming to obtain a permutation invariant representation of the input set, whilst learning a function mapping it to arbitrary target outputs \cite{Vinyals2016}. 

RPW satisfies the key property of obtaining a permutation invariant representation of its input through a variation of the attention mechanism. It can be seen as a special case of a \textit{Neural Turing Machine} \shortcite{Graves2014} or a \textit{Memory Network} \shortcite{Weston2015} in that it is a recurrent neural network model that creates a memory representation of each element in the input sequence and accesses it via an attention mechanism.

In the original RPW paper its authors also demonstrated that the order in which elements are organized as input has a significant impact on the learning performance of earlier sequence-to-sequence architectures. This is an important observation given the fact that the recurrent neural networks employed in them are, in theory, universal approximators \cite{Schafer2007}.

\subsubsection{RPW Limitations}

Whilst the RPW model constitutes a significant improvement in the way machine learning methods handle input sets, it suffers from the same limitation as Pointer Networks in terms of ordering their elements into the output sequence. Namely, it is not strictly prevented from pointing to the same element of the input set multiple times in the output, effectively returning either an invalid sequence or an incomplete reordering. This is a particularly important limitation in relation to handling \textit{multisets} (also known as \textit{msets} or \textit{bags}), where the same element can occur multiple times in the input. However, it can be mitigated through beam search or progressive masking as described in the Pointer Network section. It also suffers from a significant decrease in performance as the size of the input set increases.

\subsubsection{RPW Details} \label{rpw-details}

The RPW architecture consists of three distinct blocks:

\begin{enumerate}
    \item \textit{Read Block} - which embeds every element of the input set using the same neural network for each $x_i \in X$.
    \item \textit{Process Block} - which consists of a recurrent neural network that evolves its hidden state using a modified content-based attention mechanism to obtain a permutation invariant representation of the input over a separately predefined number of steps $t$.
    \item \textit{Write Block} - which takes the form of a Pointer Network in the set-to-sequence tasks but can also be another recurrent neural network decoder for tasks where the output elements come from a fixed dictionary.
\end{enumerate}

The Process Block evolves the permutation invariant representation of the input set by repeating the following steps $t$ times:

\begin{equation}\label{4}
\mathbf{q_t} = \textrm{LSTM}(\mathbf{q_{t-1}^*}) \qquad
\end{equation}

\begin{equation}\label{5}
z^i_{t} = f(\mathbf{m_i}, \mathbf{q_t})  \qquad \quad \
\end{equation}

\begin{equation}\label{6}
a^i_{t} = \frac{\textrm{exp}(z^i_t)}{\sum_{j} \textrm{exp}(z^j_t)}  \quad \quad \
\end{equation}

\begin{equation}\label{7}
\mathbf{r_t} = \sum_i a^i_t~\mathbf{m_i} \qquad \quad
\end{equation}

\begin{equation}\label{8}
\mathbf{q_t^*} = \langle \mathbf{q_t}, \mathbf{r_t} \rangle  \qquad \qquad
\end{equation}

where $i$ is the index over all embedded elements of the memory vector $\mathbf{m_i}$ obtained by the Read Block, $\mathbf{q_t}$ is effectively a query vector allowing us to read the permutation invariant representation $\mathbf{r_t}$ from the memories using an attention mechanism and $f()$ is any differentiable operation that takes two vectors and returns a scalar, most commonly the dot product $f(\mathbf{a}, \mathbf{b}) = \mathbf{a} \cdot \mathbf{b} = \sum_{i=1}^n a_i b_i$.

An important implementation nuance is related to the third step, where the attention vector $\mathbf{a_t}$ is obtained via a softmax operation. Depending on weights initialization, that step can result in the undefined operation of dividing infinity by infinity. This can be prevented by bounding the value range of the $\mathbf{z_t}$ vector by the use of the tanh function \cite{Logeswaran2018}.

The $\textrm{LSTM()}$ is a recurrent neural network that takes no inputs, only evolving the hidden state $\mathbf{q_t}$. The final set encoding $\mathbf{q_t^*}$, is obtained by concatenating the previous hidden state $\mathbf{q_t}$ and the permutation invariant representation $\mathbf{r_t}$. $\mathbf{q_t^*}$ becomes the hidden state input during the next iteration $t$ of the Process Block.

\subsubsection{RPW Applications}

The RPW architecture has been applied to both continuous and discrete inputs. In the former case, the input can be a set of floating point numbers or a high-dimensional embedding of the entities of interest. In the latter, it can be dictionary entries. Considerations related to the specific structure of those input elements are out of scope for this paper as they pertain to the wider topic of representation learning.

This model architecture has also been used in few-shot object recognition \shortcite{xu2017few}, graph classification \shortcite{ying2018hierarchical} and one-shot learning in the context of drug discovery \shortcite{altae2017low}. The original paper tests it on the problem of sorting a varying-size set of floating point numbers between 0 and 1. It achieves accuracy of 94\% on sets of 5 elements, performance dropping significantly for larger ones (50\% for 10 elements, 10\% for 15).

This key property of obtaining a permutation invariant representation was further formalized in the DeepSets paper, presented in the following section.

\subsection{DeepSets} \label{deepsets}

The DeepSets framework \cite{Zaheer2017} provides a robust mathematical analysis for designing \textit{permutation invariant} and \textit{permutation equivariant} deep learning models. Both of these concepts are explained in section \ref{important_concepts}. Where the \citeauthor{Vinyals2016} \citeyear{Vinyals2016}  paper focused on the former in the setting of a specific practical application, the authors of DeepSets provide a generic framework for the proper handling of set-inputs for both supervised and unsupervised learning.

The primary contribution of the DeepSets method lies in tackling the first part of the more general set-to-sequence challenge, which is to implement arbitrary set functions that result in permutation invariant representations. We have already described one example of an operation that ensured this in the RPW model's Process Block, namely a variation of the content-based attention mechanism consisting of a modified recurrent neural network. DeepSets propose a simpler, sum-based method to achieve this.

The DeepSets framework offers a simplified procedure by relying on summation of all element representations prior to further nonlinear transformations, which then transform these summed representations into the desired output (e.g. a class probability distribution or a single number for set regression). Later reimplementations of the DeepSets architecture experiment with replacing the sum operation with other permutation invariant alternatives such as taking the mean or maximum, with comparable results \cite{Lee2019}.

Additionally, the DeepSets analysis expands upon the set-input challenge by allowing for permutation equivariance, where the order of the output elements mirrors the order of the input sequence. This can be achieved, in one case, by adding a diagonal symmetry and diagonal identity constraint to the weights matrix of a fully-connected neural network layer, prior to the nonlinearity. However, set-to-sequence methods do not make extensive use of permutation equivariance therefore these are not detailed here. For more information, see the original paper.

\subsubsection{Deep Sets Details}

The proposed permutation invariant function for inference over sets takes the following general form: 

\begin{enumerate}
    \item Each element $x_i$ in the input set is transformed \textit{independently} into an embedded representation $\phi (x_i)$, possibly through multiple layers of a feed-forward neural network.
    \item The representations $\phi (x_i)$ are summed together and the result is further processed using another network $\rho$ consisting of any number of fully-connected layers with nonlinearities.
\end{enumerate}

Both $\phi$ and $\rho$ can be replaced by universal approximators, which can be learned to approximate arbitrary polynomials. In cases where additional information $q$ is available, it can be used to obtain the conditional mapping $\phi (x_i | q)$. The key to permutation invariance in this framework is simply summation of the obtained per-element representations.

\begin{equation}\label{9}
\textrm{DeepSets}(\{x_1, \ldots , x_n\}) = \rho(\textrm{sum}(\{\phi(x_1), \ldots, \phi(x_n)\})) \qquad \qquad
\end{equation}

\begin{figure}[ht]
\begin{center}
  \includegraphics[scale=0.2]{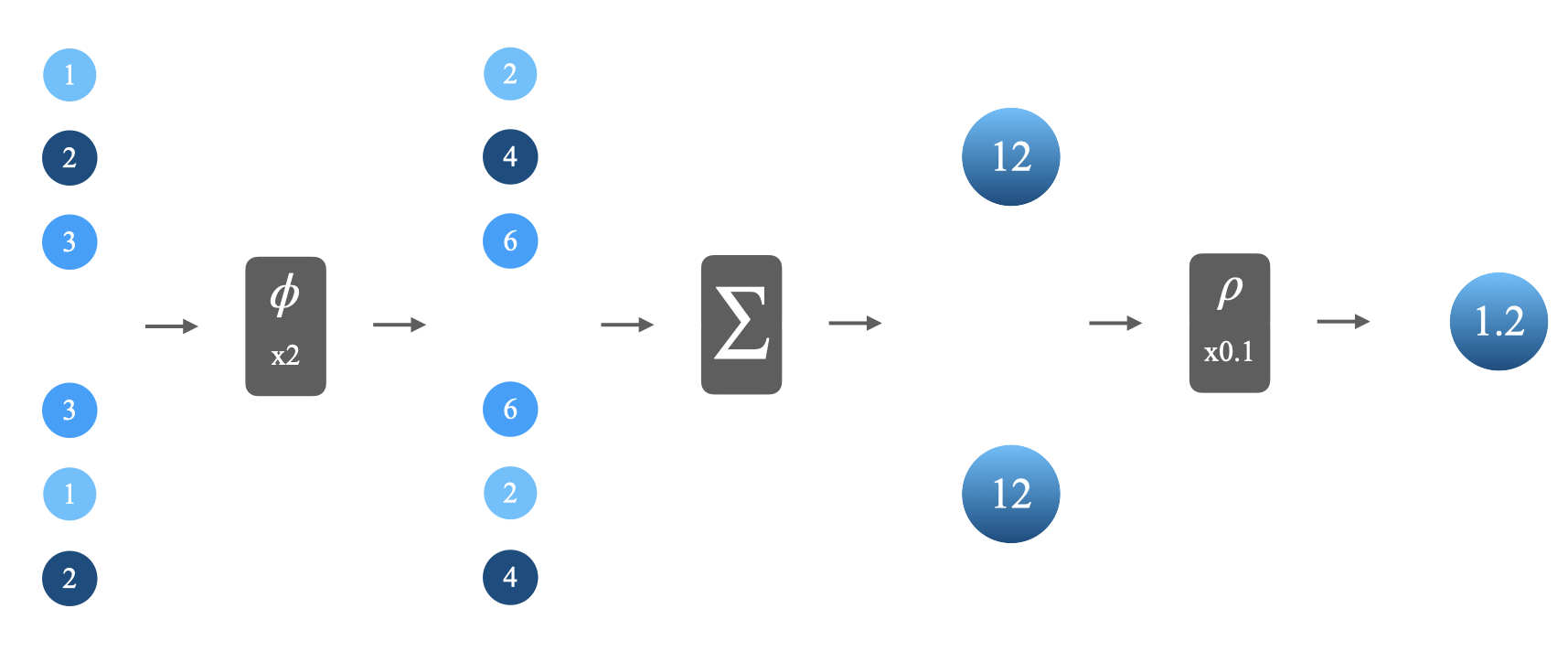}
  \caption{DeepSets}
  \label{illustration-deepsets}
  \end{center} 
  \small Every element of the two identical, shuffled sets of blue dots (leftmost) is embedded in an independent and identical way by the $\phi$ layer, resulting in a permutation equivariant transformation. These are then summed ($\Sigma$) into a permutation invariant representation and further transformed by the $\rho$ layer.
\end{figure}

\subsubsection{Deep Sets Limitations}

This approach is much simpler to implement, compared to the RNN-based Pointer Networks and the RPW model. However, it generally prevents the model from learning pair-wise and higher order interactions between the elements of the set, which are lost during the summation. Additionally, significant doubts have been raised by \shortciteauthor{Wagstaff2019} \citeyear{Wagstaff2019} relating to the limits of the representational power of the DeepSets method. More precisely, the $O(n)$ computational complexity comes at the cost of the dimensionality of the latent space having to be at least equal to the cardinality of the input set $n$ to ensure universal function approximation.

\subsubsection{Deep Sets Applications}

The DeepSets framework has been applied to point cloud classification \cite{Qi2017}, generalization tasks in reinforcement learning \shortcite{karch2020deep}, outlier detection and anomaly classification \shortcite{oladosu2020meta} among others.

\subsection{Set Transformer} \label{set-transformer}

One of the most elaborate methods designed for set-input problems is the \textit{Set Transformer} \cite{Lee2019}. This method can be seen as an extension of the popular feed-forward, attention-based Transformer \cite{Vaswani2017} to the domain of machine learning on sets. 

The Set Transformer consists of the expected stacked multi-head self-attention layers for both the internal encoder and decoder as seen in the classic Transformer. One aspect that separates it from the previously described set-to-sequence methods is that instead of using a fixed pooling operation such as summing or taking the average to ensure permutation invariance, it employs a parameterized pooling function that is learned and therefore much more adaptive to the particular task at hand. This is further referred to as \textit{Pooling by Multihead Attention} (PMA) and explained in more detail later in this section. 

The Set Transformer is specifically designed to model higher-order interactions among elements and their subsets within the input set, whilst satisfying the permutation invariance and variable input size requirements common to set-to-sequence problems. Its key novel contribution is that it concurrently encodes the entire input set through a sequence of permutation equivariant \textit{Set Attention Blocks} (SABs). By comparison, the previously discussed DeepSets method obtained element features independently of other input set elements. This modification allows the Set Transformer to  explicitly learn pairwise and even more complex interactions between set elements during the encoding step, dependent on the number of stacked SAB layers.

\subsubsection{Set Transformer Limitations}

However, the Set Transformer also introduces certain costs. The SAB, a proposed variation on the multihead attention block employed in the classic Transformer, which enables the Set Transformer to encode higher-order interactions between set elements, has the limiting quality of requiring quadratic time complexity $O(n^2)$ relative to the cardinality of the input set $n$. 

The authors of the method address this limitation by proposing an \textit{Induced Set Attention Block} (ISAB). It takes advantage of a vector of \textit{inducing points}, which is of predetermined size and is used to obtain a hidden representation of the input set by attending to it. This is in effect a low-rank projection that might be familiar to readers who have experience with autoencoder models. The technique reduces the required computation time to a linear $O(mn)$, where $m$ is the chosen number of inducing points and $n$ is the set cardinality, at the cost of reduced performance and an additional hyperparameter to be tuned.

\subsubsection{Set Transformer Details}

What follows is a more detailed inspection of the SAB, ISAB and PMA methods, which assumes some familiarity with the multihead attention mechanism proposed by \cite{Vaswani2017} as per the Transformer architecture. However, to establish full clarity, we will formally define both the basic Transformer attention and its multihead variant first.

Assume we have a set of $n$ elements as our input, each with dimensions $d_e$. First, we obtain three new vectors per each element by multiplying it with three matrices, whose weights are learned during the training step. These transformed input representations are further referred to as $n$ query vectors $Q \in \mathbb{R}^{n \times d_q}$, key vectors ($K \in \mathbb{R}^{n \times d_k}$) and value vectors ($V \in \mathbb{R}^{n \times d_v}$). These are then mapped to the desired attention outputs, applying an activation function such as softmax in the following way:

\begin{equation}\label{10}
\textrm{TransformerAttention}(Q, K, V) = \textrm{softmax} (QK^{\top})V
\end{equation}

The original implementation includes a scaling factor, which has been omitted from the above equation for the sake of simplicity. The pairwise dot product of the query $Q$ and key $K$ vectors measures how related each pair is. The final output is a weighted sum of $V$. 

The multihead attention mechanism extends this further by projecting each query, key and value onto $h$ separate vectors via sets of the three parameter matrices $W_i^Q, W_i^K, W_i^V$, one per each of the $h$ heads. Then, the TransformerAttention() function is applied to each of these $h$ vectors to obtain each head's preliminary output $O_i$:

\begin{equation}\label{11}
O_i = \textrm{TransformerAttention}(Q W_{i}^{Q}, K W_{i}^{K}, V W_{i}^{V})
\end{equation}

Finally, these outputs $\{O_i\}_{i=1}^{h}$ are concatenated and then linearly transformed:

\begin{equation}\label{12}
\textrm{MultiheadAttention}(Q, K, V) = \textrm{concatenate}(O_1, \ldots, O_h) W^0
\end{equation}

Now we can begin to move on to the Set Attention Block (SAB). It is designed to take a set and perform a slightly modified self-attention operation between its individual elements, which results in an output set of the same size. It will be useful to first define a \textit{Multihead Attention Block} (MAB), which is an intermediate building block in both SAB and its less computationally intensive alternative - ISAB. MAB takes as input two matrices of the same dimensions: $A, B \in \mathbb{R}^{n \times d}$ and first obtains a hidden representation $Z$ using a layer normalization operation as defined by \shortciteauthor{ba2016layer} \citeyear{ba2016layer}:

\begin{equation}\label{13}
Z = \textrm{LayerNormalization}(A + \textrm{MultiheadAttention}(A, B, B))
\end{equation}

Then, it processes each element in each row of $Z$ in an independent, identical way, in the same manner we have seen as part of the DeepSets method, prior to the summation. This can be done via a row-wise feed forward layer $\phi()$:

\begin{equation}\label{14}
\textrm{MAB}(A, B) = \textrm{LayerNormalization}(Z + \phi(Z))
\end{equation}

However, in our set-to-sequence setting we do not have two separate matrices of sets $A$ and $B$ that we wish to encode into some joint representation. Therefore the actual $\textrm{SAB}()$ is defined on a single matrix of an input set $X \in \mathbb{R}^{n \times d}$:

\begin{equation}\label{15}
\textrm{SAB}(X) = \textrm{MAB}(X, X)
\end{equation}

If we stack $l$ SAB layers in our set encoder, which we are able to do since both the input and output of the Set Attention Block is a set of the same size, the model's computational complexity in the cardinality of this set is $O(n^2l) \approx O(n^2)$. At the cost of quadratic computation time, stacking 2 SAB layers enables the model to encode pairwise interactions between elements. Stacking more such layers makes explicitly encoding higher-order interactions possible, which is a crucial novel contribution for tasks where such interactions define the target output.

The authors of the Set Transformer method address the computational cost of SAB by proposing a less expensive variation of it, called the Induced Set Attention Block. In ISAB, an additional array of \textit{inducing points} $I \in \mathbb{R}^{m \times d}$ is included. This vector is of predefined dimension $m$, resulting in a computational complexity $O(mn)$ or $O(lmn)$, if stacked ISAB layers are applied. The calculations within ISAB are defined as follows:

\begin{equation}\label{16}
\textrm{ISAB}_m(X) = \textrm{MAB}(X, \textrm{MAB}(I, X))
\end{equation}

The learned values of the inducing points $I$ are expected to encode large-scale aspects of the input set as meaningful features for the ultimate task. Similar to SAB, the output of ISAB is a set of the same size as the input set, which is why we still need a pooling operation to be applied to it at this point.

The final aspect of the Set Transformer that distinguishes it from the earlier set-encoding methods is the Pooling by Multihead Attention (PMA) stage. Unlike a simple sum, mean or max, the PMA pooling function has learnable parameters, which allows it to increase or decrease the relative importance given to the encoding of individual encoded elements of the output of the SAB and ISAB blocks. Usage of PMA requires specifying the number $k$ of seed vectors $S \in \mathbb{R}^{k \times d}$. Assuming we have already obtained the encoded set features $E \in \mathbb{R}^{n \times d}$ via stacked SAB or ISAB layers:

\begin{equation}\label{17}
\textrm{PMA}_k(E) = \textrm{MAB}(S, \phi(E))
\end{equation}

In most cases a single ($k=1$) seed vector is used, resulting in a single pooled set encoding, but certain clustering tasks may require multiple related outputs, justifying the use of a larger $k$.

\subsubsection{Set Transformer Applications}

The overall Set Transformer architecture can, in principle, be applied to any set-input problem (and therefore any set-to-sequence task). It will perform particularly well in problems where pairwise and higher-order interactions between the input set's elements are important to the task at hand. The authors of the original paper demonstrate the model's usefulness within such areas on the challenge of counting unique characters in a set of input images, amortized clustering with mixture of Gaussians, point cloud classification and set anomaly detection. 

It has since been applied to set-of-sets embedding problems \shortcite{Meng2019} and transfer learning in dialogue systems \shortcite{Wolf2019}.

\subsection{Other Set-Input Methods}

Below is a list of other set encoding methods that do not necessarily lend themselves directly to set-to-sequence problems, but may be of interest to the reader. The specifics of using the learned, permutation invariant set representation to produce a sequence of set elements are discussed at the beginning of the next chapter, specifically in subsection \ref{using-set-repr-to-produce-sequence}.

\subsubsection{Featurewise Sort Pooling}

This method, also known by its abbreviation as FSPool \cite{Zhang2020}, came from the field of set prediction, in relation to a problem where both the input and output can be conceived of as sets. The authors expand upon one of the naive approaches to encoding sets in a permutation invariant way. Namely, the technique of simply sorting all the elements of the input set by the values of their single chosen feature, as seen in previous work by \shortciteauthor{Zhang2018AnED} \citeyear{Zhang2018AnED}.

In set-output problems this approach results in discontinuities in the optimisation whenever two elements swap positions after the sort. This is referred to in set prediction challenges as the \textit{responsibility problem} \cite{Zhang2020}. To avoid this difficulty, the authors have developed a novel pooling method which sorts each feature across the elements of the input set and then performs a weighted sum.

This allows the model to remember the permutation applied through the featurewise sorting and apply its inverse in the decoder. This process restores the original, arbitrary order of the input elements making the encoding a permutation equivariant operation, preventing the discontinuity in the outputs of the model.

\subsubsection{Janossy Pooling} \label{janossy-pooling}

Another interesting approach to set-encoding through the use of simpler pooling operators was proposed by \citeauthor{Murphy2019} \citeyear{Murphy2019}. In the titular \textit{Janossy Pooling}, the symmetric (permutation invariant) encoding function is expressed as the average of a mixture of permutation sensitive functions applied to all reorderings of the original input.

This approach immediately raises the question of tractability. Generating all permutations of a set results in $n!$ intermediate inputs, all of which would then require the application of the chosen permutation sensitive function. To mitigate this, the authors propose a number of strategies, among them the use of a smaller number of selected canonical orderings that are presumed to carry relevant information for the specific task at hand, such as simple sorting, \textit{betweenness centrality} and others \shortcite{niepert2016learning}.

As an alternative to canonical orderings, the authors also propose a method related to a model's ability to explicitly learn pairwise and higher-order interactions between the elements of the input set. This method is referred to as $k$-ary dependencies. It consists of projecting the input to a length $k$ sequence, for example by only keeping the first $k$ elements, limiting the number of permutations that need to be averaged to $k!$, which can be tractable for a small enough $k$. The number $k$ becomes a hyperparameter capable of balancing tractability with the model's ability to learn $k$-ary interactions in the input. Finally, the authors also experiment with permutation sampling as the third method of reducing the computational complexity of Janossy Pooling, as proposed by \citeauthor{Moore2017DeepCI} \citeyear{Moore2017DeepCI} and \shortciteauthor{Hamilton2017InductiveRL} \citeyear{Hamilton2017InductiveRL} in relation to machine learning on graphs.

\subsubsection{PointNet} \label{point-net}

Not to be confused with the Pointer Network described in section \ref{pointer-networks}, \textit{PointNet} is a set-encoding method designed to handle 3D point clouds proposed by \citeauthor{Qi2017} \citeyear{Qi2017}. The geometric setting of the problem tackled by the authors of this model shares many similarities with the set-to-sequence domain. These include the lack of order in the input, requiring the use of symmetric, permutation invariant encoding functions, and the importance of interactions between individual input elements or, specifically in this case, points. An additional requirement for this setting is invariance under certain geometric transformations of the entire point cloud. For example, if we are tasked with classifying 3D objects represented by a point cloud, we want to correctly classify a chair regardless of its rotation and translation.

In practice, PointNet first obtains an embedding of each of the input points through stacked, fully-connected layers in the form of a multilayer perceptron, such that each element is identically and independently transformed. This permutation equivariant representation is then pooled via the $\textrm{max}()$ operator (per dimension) and further transformed through an additional fully-connected layer. This is in effect a slight variation of the procedure proposed by \citeauthor{Zaheer2017} \citeyear{Zaheer2017} and discussed in section \ref{deepsets}. Other pooling operations, including an attention-based sum inspired by the RPW model \cite{Vinyals2016}, are also experimented with in the original paper.. 

Finally, the obtained point cloud encoding is concatenated with the embedding of each point, reminiscent of the approach seen in listwise ranking, described in section \ref{listwise-ranking}.
This combination of local and global features is shown to be crucial for point segmentation tasks. The authors also provide proof that their network is a universal approximator for continuous set functions and demonstrate its robustness to small perturbations of the input set.

\subsubsection{AttSets} \label{att-sets}

The \textit{AttSets} model, proposed by \citeauthor{yang2020robust} \citeyear{yang2020robust}, uses weighted attention to obtain a permutation invariant representation of the input set. It was originally applied to a multi-view 3D reconstruction task, where a set of images of the same object from different angles is used to estimate its true three dimensional shape.

AttSets improves the performance of previous, simpler pooling functions used for 3D object recognition. These include both first-order operators such as max(), average() and sum(), which do not have any trainable parameters, as well as higher-order statistical functions such as bilinear pooling \shortcite{lin2018second}, log-covariance pooling \shortcite{ionescu2015matrix} and harmonized bilinear pooling \shortcite{yu2018multi}, which have only few. 

In order to achieve this, each element of the set is individually and independently transformed via a learned attention function, which can take the form of a fully connected layer or a multidimensional convolutional layer, depending on the form of the input. The output of this function is normalized via softmax() and then used as an attention mask over the original input elements. This allows the model to learn to pay a varying degree of attention to individual dimensions of the input elements' representations. To obtain the final, fixed-length set encoding, the original input elements are multiplied by the attention mask and summed together.

\subsubsection{RepSet} \label{rep-set}

An interesting set-encoding method, referred to as \textit{RepSet}, has been proposed by \citeauthor{Skianis2020} \citeyear{Skianis2020}. The RepSet model consists of stacked feed-forward, fully connected layers, reminiscent of the DeepSets method \cite{Zaheer2017}, followed by a custom permutation invariant layer replacing the sum() operator. This layer is inspired by concepts from the field of bipartite graph matching and has allowed the model to show promising performance on text and graph classification tasks.

The permutation invariance is achieved through a configurable number of hidden sets (potentially of different cardinalities), whose elements correspond to columns of trainable weight matrices. These are then compared with the elements of the actual input set to create matrices that are fed to a bipartite matching algorithm, specifically the Hungarian Algorithm \cite{grinman2015hungarian}. The resulting values can be further transformed through standard neural network layers for set classification and regression purposes.

A significant issue with this approach is the computational complexity $O(mn + n^2~\textrm{log}~n)$, where $n$ is the cardinality of the input set and $m$ is the chosen number of hidden sets. This characteristic, limiting the usefulness of the method regarding larger sets, stems from the bipartite matching algorithm needed to obtain the final set encoding. The authors of the method address this by proposing a relaxation of RepSet, referred to as \textit{ApproxRepSet} \cite{Skianis2020}, which removes one of the constraints on the range of values taken by the elements of the hidden sets.

\section{Ordering Methods} \label{orderingmethods}

This section focuses on the second of the two primary challenges inherent to set-to-sequence tasks, which is outputting a permutation. Assuming we are able to obtain a meaningful representation of an input set of any length, how do we use that representation to produce a reordering of the input's original elements? This is a constraint that is easy to satisfy when designing traditional combinatorial optimization algorithms, yet in deep learning it requires relatively complex model architectures \cite{Bengio2020}.

Three particular branches of ordering methods have emerged as most prominent in deep learning:

\begin{enumerate}
    \item \textit{Attention-based Pointing}
    
    In which a vector of attention weights over all elements of the input set is generated iteratively at every index of the output sequence. The highest attention value points to the element that should be placed in the current position within the sequence.
    
    \itemsep0.5em 
    \item \textit{Permutation Matrices}
    
    Where a square, binary, doubly-stochastic matrix or a relaxation thereof is generated for each input set. The index of the highest value in each row identifies the element that should take the position at the same index as the number of the row. The input can be left-multiplied by this matrix to obtain the final reordering.
    
    \itemsep0.5em 
    \item \textit{Listwise Ranking}
    
    In ranking methods the target order is represented through the assignment of a score to each element of the input set, which enables the final permutation to be obtained through sorting. Listwise ranking takes into consideration the relative scores of all other set elements when computing the score for a particular one.
    
\end{enumerate}

Each of these three basic frameworks is described in the following sections, with references to specific methods that expand upon them, where relevant. First, however, we must discuss how best to utilize the permutation invariant set representation, obtained via the set encoding methods discussed in section \ref{set-encoding-methods}, to output the target sequence.

\subsection{From Set Representation to a Sequence}
\label{using-set-repr-to-produce-sequence}

As stated previously, deep learning models do not directly take unordered sets as input. Instead they transform ordered arrays representing one (usually arbitrary) of $n!$ permutations of a given set's elements. Therefore, an inductive bias is introduced into the model's internal architecture to first obtain a permutation invariant representation of the underlying input set, which by definition will be the same regardless of which arbitrary permutation the model happened to receive.

However, when our prediction target is the optimal order of a given set's elements and we are directly feeding our model an array in some arbitrary order, then our target output sequence must refer to this initial, random order when predicting the preferred one. In effect, in the case of simple permuted sequences without repetition and exclusion, we are seeking a permutation equivariant function that outputs ranks, conditioned on a permutation invariant representation of the entire set. In the case of complex permutations, we also expect the same sequence to be predicted based on the same input set. 

Regardless, the output sequence has to refer to the arbitrary order in which the input set elements are presented to it. In essence, whilst we assume there is one optimal order of a given set's elements, there are $n!$ target permutations relative to the $n!$ arbitrary ones that form the actual input of the model, per input set. This is because each target permutation reorders an arbitrary one that the model receives as actual input, each representing the same, singular optimal order of the underlying set's elements.

The remaining question is how to utilize the fixed-length, permutation invariant set representation when outputting the sequence representing a permutation of the arbitrarily ordered input's elements. There are two dominant approaches, namely summation and concatenation of the representation of the entire set with the learned representations of each element. In this way, every transformation of individual set elements has the chance to take into consideration the entire available set, as previously seen in the PointNet model by \citeauthor{Qi2017} \citeyear{Qi2017}. 

Considerations regarding which of those two methods is preferable relate to the larger field of representation learning, with ample examples both on the side of summation \shortcite{szegedy2016rethinking_inception} and concatenation \shortcite{noreen2020deep_concatenation}. These include the \textit{ResNet} architecture by \shortciteauthor{resnet_he2016deep} \citeyear{resnet_he2016deep}, with addition used in the eponymous residual connections, and the \textit{DenseNet} model and its descendants, employing concatenation \shortcite{huang2018condensenet}, within the field of computer vision.

An alternative approach, applicable to RNN-based ordering methods, would be to utilize the set representation in the hidden state of the recurrent network. Similarly, attention-based methods could use the set representation to influence which elements the model focuses on. Finally, in ranking approaches the set representation can rather elegantly be used in place of the \textit{query} (see section \ref{listwise-ranking} for more details). Investigation of the efficacy of these methods is a possible direction for future research, as outlined in section \ref{furtherresearch}.

\subsection{Attention-Based Pointing} \label{attention-based-pointing}

The term \textit{attention} covers a wide spectrum of methods within the field of machine learning. First introduced as part of an encoder-decoder model applied to a sequence-to-sequence translation task by \citeauthor{bahdanau2014neural} \citeyear{bahdanau2014neural}, it has since been utilized in a number of other domains, including computer vision \shortcite{xu2015show}, graph-based problems \shortcite{velickovic2018try}, reinforcement learning \cite{iqbal2019actor} and many more. For a comprehensive overview, see \shortciteauthor{chaudhari2019attentive} \citeyear{chaudhari2019attentive}. 

In structured output tasks that are inherent to the set-to-sequence domain, the key idea is to induce a vector of attention weights over the entire input, allowing for the selection of elements by their position, conditioned on the learned representation of the entire input and the sequentially predicted outputs. An example of such a setting from the field of natural language processing (NLP) is \textit{linearization}, which involves solving a problem that will be familiar to most readers who have tried to learn a new language. Namely, given a randomly shuffled set of words, we are tasked with ordering them into a properly structured sentence. A survey of these and other attention-based methods in NLP can be found in the work of \citeauthor{hu2019introductory} \citeyear{hu2019introductory}. 

In effect, the model can learn to identify the current most pertinent position in the input through the attention weights, which we then interpret as pointers to the next preferred element for the output sequence. It may be worth mentioning that valid concerns have been raised as to the overall interpretability of attention weights in text classification and question answering tasks by \citeauthor{jain2019attention} \citeyear{jain2019attention}. However, in attention-based pointing these weights have a direct impact on the predicted permutation, leaving little room for interpretational ambiguity.

\subsubsection{Details of Attention-Based Pointing}

Attention-based pointing is an adaptation of content-based attention \cite{Graves2014} to an ordering challenge. The model learns a distribution over all input elements at each position of the output sequence. This should not be confused with self-attention, also known as intra-attention \shortcite{cheng-etal-2016-long}, where the target sequence is always the same as the input. In self-attention, the parameterized attention function learns to relate different positions of the input sequence to each other. In attention-based pointing the attention function learns to relate the elements of the input with the current positions of the output.

Assume we have a randomly ordered input sequence of varying length $n$, consisting of $d$-dimensional elements: $X = \langle \mathbf{x_1}, \ldots, \mathbf{x_n}\rangle$ such that $X \in \mathbb{R}^{n\times d}$, which we want to order according to some preference. This sequence is an $n$-tuple, representing a single permutation of the corresponding set $X^\prime = \{\mathbf{x_1}, \ldots, \mathbf{x_n}\}$, which we cannot represent directly as a single vector, due to the fact that vectors posses inherent order whereas sets do not. 

Our target output is a sequence of integer pointers $Y = \langle y_1, \ldots, y_n : y_i \in \mathbb{Z^+} \rangle, $, satisfying the conditions: $\forall i, j: y_i \neq y_j$ and $\forall i, 1 \leq y_i \leq n$, signifying that no element from the input can take multiple positions in the output sequence and every element must be assigned a position in the final reordering. Our objective is to define a differentiable function $f : X \rightarrow Y$ that meets these criteria, with some parameters $\theta$, lending itself to gradient-based training.

The simplest effective formulation of such an attention-based ordering function requires the following learnable parameters: two square weights matrices $W_1, W_2 \in \mathbb{R}^{d \times d}$ as well as a single vector $\mathbf{v} \in \mathbb{R}^{d \times 1}$. Additionally, given the sequential nature in which the ordering is generated, we also have access to a decoder state $\mathbf{s_t}$ at each step $t$, for which we calculate the attention-based probability distribution over input elements. The array of encoded set elements $X$ is processed in the following way:

\begin{equation}\label{18}
z^i_t = \mathbf{x_i} W_1 + \mathbf{s_t} W_2 \ \ \ \ \ \quad \qquad \textrm{for}~i \in (1, \ldots ,n)
\end{equation}

\begin{equation}\label{19}
a^i_t = \frac{\textrm{exp}\left(\mathbf{v}^{T} \sigma\left(z^i_t\right)\right)}{\sum_{j}^{n} \textrm{exp}\left(\mathbf{v}^{T} \sigma\left(z^j_t\right)\right)} \quad \ \textrm{for}~i \in (1, \ldots ,n)
\end{equation}

\begin{equation}\label{20}
y_t = \textrm{argmax}(\mathbf{a_t}) \ \quad \qquad \qquad \ \textrm{for}~t \in (1, \ldots ,n)
\end{equation}

where $\sigma$ is a nonlinear activation function, commonly the hyperbolic tangent \cite{Logeswaran2018}. The vector $\mathbf{z_t} = \langle z^i_t, \ldots, z^n_t \rangle$ is a representation of all input elements, adjusted by the representation of the generated output sequence at step $t$, through additive attention \cite{bahdanau2014neural}. Finally, $y_t$ is an integer pointer to the element of $X$ which received the highest attention value within the attention vector $\mathbf{a_t} = \langle a^i_t, \ldots, a^i_n\rangle$ at the current step $t$ of the sequentially predicted output sequence.

Most methods obtain the embedding of the entire input $X$ through more complex encoding methods, such as stacked bidirectional LSTMs, as seen in the work of \citeauthor{Vinyals2015} \citeyear{Vinyals2015} and \citeauthor{Vinyals2016} \citeyear{Vinyals2016}. This representation retains the information about the original arbitrary order of the elements and should be dependent on the entire set. In the above example, the length $t$ of the output sequence of pointers $Y$  is the same as the cardinality of $X^\prime$, which is the case in many combinatorial optimization problems. However, in some structured prediction tasks the optimal length of the output is also subject to the learning process, allowing for pointing to the same element $\mathbf{x_i}$ multiple times or not at all, as is the case in catalog design \shortcite{CarlsonSkalak1998UsingAE}.

It should be noted that in the above formulation nothing is explicitly preventing the model from pointing to the same element $\mathbf{x_i}$ at multiple positions of the output sequence. In practice the iterative learning process can largely prevent this from occurring, given appropriate training data. One method to explicitly prohibit repetition is progressive masking, as described in section \ref{pointer-networks-limitations}. 

In practice, beam search is commonly used during inference to increase the probability that the most optimal sequence is predicted, as evidenced in the work of \citeauthor{Bahdanau2017RL} \citeyear{Bahdanau2017RL} and \citeauthor{Kool2019} \citeyear{Kool2019}. Beam search employs a heuristic search algorithm to expand a limited number of most promising vertices of the graph defined by the attention vectors over a predefined number of output steps. It tracks a small number of potential partial solutions, at the controlled cost of memory and computation. However, the exact reason why a small beam number results in qualitatively better predictions has recently been called into question \shortcite{meister2020if}. For a more detailed overview in the context of sequence prediction, see \citeauthor{wiseman-rush-2016-sequence} \citeyear{wiseman-rush-2016-sequence}.

\subsubsection{Methods Using Attention-Based Pointing}

Many neural network model architectures across distant fields of application include variations on the basic attention-based pointing mechanism. These include specifically set-to-sequence methods, such as the Pointer Networks discussed in section \ref{pointer-networks-details} and the RPW model detailed in section \ref{rpw-details}, as well as elements of complex models in reinforcement learning applied to competitive real-time strategy challenges \cite{Vinyals2019}. 

Additionally, attention-based pointing found usage in identifying entailment between documents \shortcite{Rocktschel2016ReasoningAE}, abstractive text summarization \shortcite{see2017get}, rare and out-of-vocabulary word prediction \shortcite{merity2016pointer} as well as describing multimedia content \shortcite{cho2015ieee}. An interesting augmentation of attention-based pointing has been experimented with in the context of generating structured queries (in SQL) from natural language sequences \shortcite{zhong2018seq2sql}. However, the improvement stems from augmenting the input with SQL keywords to limit the output space, not from an essential adjustment to the underlying attention-based pointing mechanism. 

An application of attention-based pointing to generate solutions to another classic combinatorial optimization challenge, the Vehicle Routing Problem, was proposed by \citeauthor{Kool2019} \citeyear{Kool2019}. The resulting architecture is an encoder-decoder \textit{Graph Attention Network} \cite{velickovic2018try}, employing multiheaded attention in the encoder and node masking in the decoder, which uses the embeddings of all the nodes and the entire graph at each step $t$ to point to the next node to be visited. Unlike the previously discussed methods, this model is trained using a gradient estimator from the field of reinforcement learning, first proposed by \citeauthor{williams1992simple} \citeyear{williams1992simple}.

\subsection{Permutation Matrices}\label{permutation-matrices}

A permutation matrix is a square, binary matrix $P$ having exactly one entry $p^i_j = 1$ in each row $i$ and each column $j$ \cite{STUART1991255}. All other entries are equal to $0$. $P$'s dimensions are defined by the length of the desired output sequence, most commonly equal to the length of the input. A permutation matrix is unimodal in that each of its rows has a unique, highest value. It is also doubly-stochastic, since it consists entirely of nonnegative numbers, with each row and column summing to 1.

The key property of a permutation matrix is that given an arbitrarily ordered sequence, we can left-multiply it by a permutation matrix to obtain a reordered sequence. For example, given an arbitrarily shuffled input sequence $\mathbf{x} = \langle x_1, x_5, x_3, x_4, x_2 \rangle$, which we want to permute in such a way as to restore the $i-$indexed ascending order to obtain $\mathbf{x^{\pi}} = \langle x_1, x_2, x_3, x_4, x_5 \rangle$, we can predict the permutation matrix visualized below. Our target is the permutation $\pi$, given in two-line notation:

\begin{equation}\label{21}
\pi = \begin{pmatrix}
x_1 & x_5 & x_3 & x_4 & x_2\\
1 & 5 & 3 & 4 & 2
\end{pmatrix}
\end{equation}

We can apply this permutation to the transposed input $\mathbf{x^{\top}}$ via left-multiplication by the target permutation matrix $P$:

\begin{equation}\label{22}
\mathbf{x^{\pi}} = P \mathbf{x^{\top}} = \begin{bmatrix}
1 & 0 & 0 & 0 & 0\\
0 & 0 & 0 & 0 & 1\\
0 & 0 & 1 & 0 & 0\\
0 & 0 & 0 & 1 & 0\\
0 & 1 & 0 & 0 & 0
\end{bmatrix} \begin{bmatrix}
x_1\\
x_5\\
x_3\\
x_4\\
x_2
\end{bmatrix} = 
\begin{bmatrix}
x_1\\
x_2\\
x_3\\
x_4\\
x_5
\end{bmatrix}
\end{equation}

This gives us a way to represent the target permutation in the form of a matrix of numbers between 0 and 1, marking an important step towards enabling the application of machine learning methods.

\subsubsection{Making Permutation Matrices Learnable}

In order to use a permutation matrix as the target output of a machine learning model that can be trained through some form of gradient-based iterative optimization, we must introduce certain relaxations of the concept. Otherwise, we are left with the equivalent of a sorting operator which is non-differentiable \shortcite{grover2019stochastic}.

Traditionally, sorting operations result in either a permutation $\pi = \langle 1, 5, 3, 4, 2 \rangle$ (from the previous example) or a vector of reordered elements $\mathbf{x^{\pi}} = \langle x_1, x_2, x_3, x_4, x_5 \rangle$. Much like the permutation matrix, the former is non-differentiable with respect to the input due to being integer-valued and the latter due to being piecewise linear \shortcite{cuturi2019differentiable}.

The most direct way to obtain a differentiable relaxation $P^{\prime}$ of the permutation matrix $P$ is to map the input $\mathbf{x}$ to a continuous codomain, as opposed to the original discrete one. Thus, we need a relaxation such that $P^{\prime} \in \mathbb{R}^{n \times n}$, lending itself to gradient-based optimization. Additionally, an efficient projection from the continuous codomain back to the discrete one must exist to allow for the use of loss functions and evaluation metrics. This can be achieved by applying the argmax() function per row of $P^{\prime}$ to find the position in which the unique $1$ would have been located in the actual permutation matrix $P$. 

The resulting relaxation must retain the property of row-stochasticity, such that:

\begin{equation}\label{23}
\forall i, j \in \{1, \ldots, n\}: p^{\prime}_{ij} \geq 0
\end{equation}

\begin{equation}\label{24}
\forall i \in \{1, \ldots, n\}:
\sum_{j=1}^{n} p^{\prime}_{ij} = 1
\end{equation}

and of unimodality, such that a vector $\mathbf{y}$ is both obtainable from the relaxation matrix $P^{\prime}$ through row-wise argmax() and a valid permutation of the input:

\begin{equation}\label{25}
\forall i \in \{1, \ldots, n\}: y_i = \textrm{argmax}(
\mathbf{p}^{\prime}_i)
\end{equation}

There are many possible methods of obtaining this relaxation from the input after it has been transformed by a chosen neural network architecture, such as adding elementwise Gumbel perturbations \shortcite{Mena2018}, applying the Sinkhorn operator to directly sample matrices near the Birkhoff polytope \shortcite{linderman2018reparameterizing}, which is the convex hull whose points are doubly-stochastic matrices \cite{Emami2018LearningPW}, or through the application of a softmax() operator on a derived matrix of absolute pairwise distances between the individual input elements \cite{grover2019stochastic}.

Depending on the specifics of the task at hand, the target matrix can be predicted in a single pass (if the length of the input and output is a constant, known prior to inference) or sequentially, row by row \cite{Nishida2017}. An interesting method employing permutation matrices has been proposed by \shortciteauthor{Zhang2019permcomp} \citeyear{Zhang2019permcomp}, in which a trainable, pairwise \textit{ordering cost} function is used to produce an anti-symmetric matrix $C$, whose entry $c_{ij}$  represents the cost of placing the $i$-th element before the $j$-th. This function is represented as a neural network, which is then used to continuously adjust the learned permutation matrix. This is referred to as a \textit{Permutation-Optimisation} module, and has been demonstrated to perform well on number sorting, re-assembling image mosaics and visual question answering, with one limiting feature of entailing cubic time complexity.

\subsection{Listwise Ranking} \label{listwise-ranking}

Ranking methods determine the predicted order of elements by assigning a score to each element and then sorting them according to these scores. In listwise ranking, a score is calculated based on a list of available elements and a \textit{query}, for which a specific order is to be predicted. The terminology stems from applications in information retrieval, where the task is to rank the available \textit{documents} (such as web pages) in order of relevance for a given query (e.g. a search term). For an overview of neural ranking methods, see \shortciteauthor{mitra2018introduction} \citeyear{mitra2018introduction}.

Listwise ranking is distinguished from point- and pair-wise ranking. Pointwise methods reduce the ranking problem to regression, in that the relevance score for an element is obtained only from its own representation and the query. Pairwise approaches reduce it to binary classification \shortcite{lei2017alternating}. Given a query and a pair of elements, they predict which element is the more relevant of the two. In listwise ranking, the prediction is performed on a list of objects.

Application of listwise ranking methods to set-to-sequence problems requires certain adjustments. Most importantly, in such ordering and structure prediction challenges we are not given a specific query, for which an ordering is to be generated. However, we are able to obtain a permutation invariant set representation through the set encoding methods detailed in section \ref{set-encoding-methods}. We can use this learned embedding to fill the role that a query performs in traditional learn-to-rank methods. Intuitively, the relative rank of each element in the output sequence should depend on the entire available set.

This marks an important departure from the assumption that two elements have a canonical relative order, which should remain unchanged regardless of what other elements are present in the input set. In practice, higher-order interactions between available elements can entirely change the target sequence. In order to learn such properties, a ranking method can utilize the permutation invariant set representation as the query and predict the relevance scores of all available elements in one go, in the listwise manner described in the following subsection.

\subsubsection{Details of Listwise Ranking} \label{details-of-listwise-ranking}

In order to illustrate the basic underlying mechanisms in listwise ranking methods, this subsection investigates the first method that formulated order prediction on a list of objects, namely the \textit{ListNet} model, proposed by \cite{cao2007learning}. From this point on, whenever we refer to a query, we are referring to the permutation invariant representation of the entire set obtained via the previously outlined set encoding methods. This representation can be learned in parallel with the weights of the ranking the model.

Once more, assume we are given an arbitrarily ordered input sequence $X = \langle \mathbf{x_1}, \ldots, \mathbf{x_n}\rangle$ such that $X \in \mathbb{R}^{n\times d}$. This sequence is one of $n!$ possible permutations of the corresponding set $X^\prime = \{\mathbf{x_1}, \ldots, \mathbf{x_n}\}$. The objective is to transform the given $X$ vector into the best possible permutation of $X^{\prime}$, according to some preference. This target permutation is $\mathbf{\pi} = \langle \pi(1), \ldots, \pi(n) \rangle$, such that $\pi(i)$ is the object at position $i$ in the permutation. The set of all possible permutations of length $n$ is denoted as $\Omega_n$. The target permutation $\pi$ is represented by a vector of scores $\mathbf{y^\pi} = \langle y_{\pi(i)}, \ldots, y_{\pi(n)} \rangle$, where $y_{\pi(i)}$ is the predicted relevance score for the element $\mathbf{x_i}$. All elements of the original input $X$ get assigned a score by the learned neural network, in a listwise manner.

A naive approach would be to try to obtain the probability $P_y(\pi)$ of each of the $n!$ possible permutations from $\Omega_n$, given a set of scores $\mathbf{y}$:

\begin{equation}\label{26}
P_y(\pi) = \prod_{i=1}^{n} \frac{\phi(y_{\pi(i)})}{\sum_{k=i}^{n} \phi(y_{\pi(k)})}
\end{equation}

where $\phi()$ is any increasing and strictly positive function applied to the scores and $y_{\pi(i)}$ is the relevance score of element $i$ in the vector $\mathbf{y_\pi}$ corresponding to some permutation $\pi$ \cite{cao2007learning}. 

However, this is a computationally inefficient approach, therefore we instead calculate the \textit{top one probability} for each element $\mathbf{x_i}$. This probability is equal to the sum of the permutation probabilities of every permutation where the $i$-th element was ranked first. For the exact proof of this equivalence, see the appendix in \citeauthor{cao2007learning} \citeyear{cao2007learning}. The crucial observation is that we do not need to calculate all the permutation probabilities to obtain the correct top one probability $P_{top}(\mathbf{x_i})$ of each element, given a list of scores $\mathbf{y_\pi}$:

\begin{equation}\label{27}
P_{top}\left(\mathbf{x_i}\right) = \frac{\textrm{exp}\left(y_{\pi\left(i\right)}\right)}{\sum_{k=1}^{n} \textrm{exp}\left(y_{\pi\left(k\right)}\right)}
\end{equation}

In effect, the probability distribution over permutations is obtained by applying the softmax() function to the predicted relevance scores. Since this is a supervised learning framework, our model has access to the ground truth distribution for each example list. This allows for the use of loss functions that compare two distributions, such as the Kullback-Leibler (KL) divergence, also referred to as relative entropy \cite{liu2003kullback}.

In order to obtain the vector of relevance scores $\mathbf{y_\pi}$, the query vector $\mathbf{q} \in \mathbb{R}^m$ is concatenated with a fixed-length, learned embedding of each element of the input set $\mathbf{x_i}$, resulting in the final input $I \in \mathbb{R}^{n \times (m + d)}$, per single example list. In the original ListNet model, this input is transformed through stacked, fully-connected layers, of which the last one has a single unit outputting the relevance score per element. 

As such, this transforms each concatenation of the query and the embedded element in an independent and identical manner. This is practical for ranking tasks, where two items always have the same relative target order, given a query. However, in more complex structure prediction tasks, common in set-to-sequence challenges, relying entirely on the set encoding $\mathbf{q}$ and the listwise loss to identify interactions between set elements limits the ability to approximate complex functions, in which higher-order interactions between elements have an impact on the final order \shortcite{lan2009generalization}. This limitations of ListNet is partially addressed by later methods, discussed in section \ref{methods-using-listwise-ranking}.

Additionally, a limitation of the ranking approach is that by representing the target output sequence through relevance scores per input element, we cannot learn to output sequences of any length other than $n$. This also precludes the use of listwise ranking in challenges where the output sequence is of the same length as the number of input elements $n$, but repetition and exclusion of elements is required. For a more detailed look into the underlying theoretical aspects of listwise ranking, see \shortciteauthor{xia2008listwise} \citeyear{xia2008listwise}.

\subsubsection{Methods Using Listwise Ranking} \label{methods-using-listwise-ranking}

Multiple listwise ranking methods in machine learning have been developed since the original ListNet model. A notable framework that addresses the challenge of learning pairwise interactions between elements came in the form of \textit{BoltzRank}, in which the rank probabilities are sampled from a Boltzmann distribution, employing an energy function that depends on a score influenced by both individual and pairwise potentials \cite{volkovs2009boltzrank}.

Another approach of interest is the \textit{FATE} framework proposed by \shortciteauthor{Pfannschmidt2018DeepAF} \citeyear{Pfannschmidt2018DeepAF}. The authors identify the problem of predicting a relevance score for an element and the query with only the loss function carrying the signal regarding the context of other available elements. In order to address this, they effectively employ a permutation invariant set-encoder \cite{Zaheer2017}, whose output is concatenated to the learned representation of each element, in a variation of the basic method described in section \ref{details-of-listwise-ranking}.

\shortciteauthor{ai2018learning} \citeyear{ai2018learning} propose a complex architecture combining two-stage ranking, sequential recurrent neural networks and an attention-based loss function. The proposed \textit{Deep Listwise Context Model} (DLCM) sequentially encodes the most relevant results using the corresponding feature vectors, trains an additional, local context model and employs it to re-rank the best $k$ results. 

Finally, \shortciteauthor{pang2020setrank} \citeyear{pang2020setrank} employ the Set Transformer to obtain element representations that encode cross-document interactions and return a permutation invariant ranking by sorting the permutation equivariant relevance scores per each document.

\section{Discussion} \label{discussion}

In this section, the key challenges and contexts of application for set-to-sequence deep learning methods are discussed. Additionally, separate sections are devoted to the progress in set encoding and permutation learning, along with current limitations and proposed directions for future research.

\subsection{Key Challenges}

Core challenges of set-to-sequence include representing sets of varying sizes in a permutation invariant way, encoding pairwise and higher-order interactions between set elements and keeping the computational complexity moderate in the cardinality of the input set. This last requirement stems from the fact that in many important set-to-sequence problems the size of the input set can be quite large, particularly in computer vision point-cloud based tasks \shortcite{ge2018hand}. This has been addressed by various proposed set-encoding or set-pooling methods discussed in section \ref{set-encoding-methods}, such as RNN-based Pointer Networks \cite{Vinyals2015} and Read-Process-and-Write models \cite{Vinyals2016} as well as through fully connected methods such as DeepSets \cite{Zaheer2017} and the Set Transformer \cite{Lee2019}, among others.

Another set of key challenges in set-to-sequence stems from the many possible ways of formalizing the process of outputting a permutation sequence. Potential methods include learn-to-rank approaches, permutation matrices and attention-based methods discussed in section \ref{orderingmethods}. Additionally, it is not always trivial to combine the output of the specific set-encoding technique with the expected input of these permutation methods in a way that results in good model performance. Finally, we expect not just the set-encoding module but the entire set-to-sequence model to be permutation invariant and always give the optimal output regardless of how the array representing the set is reordered at input. 

\subsection{Contexts of Application}

There are many contexts where the input data does not have an inherent ordering and the number of input elements varies (i.e. \textit{set-input} problems) and possibly even more where the input elements are not unique, instead repeating a meaningful number of times, in which case the task presents an \textit{mset-} or \textit{multiset-input} problem. Additionally, the set elements may be allowed to reoccur multiple times in the predicted sequence or be excluded from it entirely, forming a \textit{complex permutation}. These elements can be both continuous (e.g. word embeddings) or discrete (coming from an index-based dictionary). An example of such a problem with a strong industrial application is the question of how to order available product offers into a displayable catalogue that will keep the reader engaged and eventually inspire them to make a relevant purchase \cite{Liao2004}. 

Similar challenges are faced by the experts who order news articles into a coherent publication, book authors composing chapters into a novel and by engineers tackling the challenge of \textit{catalog design}, in which a
configuration is created by assembling off-the-shelf components into a functional system \cite{CarlsonSkalak1998UsingAE}. Set-to-sequence methods can also benefit architects in their search for the best configurations of the design space, taking into consideration structural efficiency, daylight availability and other aspects of building performance \cite{Brown2017}. Whenever the elements come from a pre-existing set and the output is structured as a complex permutation, we are facing a set-to-sequence challenge.

\subsection{Progress in Set Encoding}

As machine learning methods become more and more widely used, the range of input and output data structures that these methods are applied to becomes larger. Mirroring the development and consecutive growth in popularity of methods created specifically to handle certain types of input data, such as convolutional neural networks for images and recurrent models for sequences, one of the most significant developments in set-to-sequence has been the work towards obtaining permutation invariant representations of sets of varying lengths.

Particularly, this representation can now be obtained in a way that is capable of explicitly encoding not just pairwise but also higher-order interactions between input elements. Intuitively these improvements mirror the way we as humans process input sets, in that we group the elements into meaningful collections, allowing for enough flexibility that this grouping may change entirely upon the introduction of a new element.

\subsection{Progress in Permutation Learning}

Comparative progress has been made regarding the ways in which we formalize the permutation tasks at hand. Given that the simplest form of permutation (reordering) is sorting, the early methods formalized this challenge as a ranking problem, with pointwise, pairwise and eventually listwise loss being used to train the model. These methods suffer from the assumption that any two elements have a canonical pairwise ranking, regardless of the features of the other elements in the input set.

Another alternative emerged in the form of self-attention applied in such a way as to output softmaxed pointers to the elements of the input set at each step in the output sequence, with the additional use of beam search during inference \cite{Vinyals2015}. The third dominant formalization comes in the form of learning to output a permutation matrix. This allows for the original input to then be matrix-multiplied by the row-stochastic permutation matrix, resulting in the desired reordering \cite{Nishida2017}, \cite{Emami2018LearningPW}. It is a sign of the complexity of the field that no clear preferred formalization of its core challenge has emerged, with ranking (for example) still finding useful application in active research \shortcite{Kumar2020}.

\subsection{Limitations}

A crucial limitation of many of the cutting edge set-to-sequence models, such as the Set Transformer \cite{Lee2019} and the permutation mechanism in Pointer Networks, is their reliance on self-attention. Whilst Transformer-based methods that rely solely on multiheaded self-attention have seen remarkable success, even in applications beyond a fixed-length context \shortcite{dai-etal-2019-transformer}, their ability to process hierarchical structure has hard limits. Specifically, purely self-attention architectures are entirely dependent on the number of multi-attention heads and layers growing alongside the size of the input to retain the ability to model recursion, finite-state languages and other hierarchical aspects of our data \cite{Hahn2020}.

In order to overcome this limitation, efforts have been made to combine the benefits of sequential computation, inherent to recurrent neural networks, with the advantages of parallel computation and the global receptive field of the Transformer. This method is referred to as the \textit{Universal Transformer} \shortcite{Dehghani2019}. In addition, it also includes the \textit{Adaptive Computation Time} mechanism proposed by \citeauthor{Graves2016-adaptive-time} \citeyear{Graves2016-adaptive-time}, which enables the model to dynamically learn how many computational steps to perform per the features of the input sequence. However, these advances have not yet been translated to the domain of set-input challenges. 

\subsection{Directions for Further Research} \label{furtherresearch}

There are many possible directions for further investigations, pertaining to both the area of encoding sets and proposing novel permutation learning methods. As hinted in section \ref{using-set-repr-to-produce-sequence}, the question of how best to utilize the fixed length set representation within the internals of the ordering module remains open. However, we believe the areas of learning complex permutations (\ref{complex-permutations}) and differentiable loss functions (\ref{differentiable-loss-functions}) deserve separate attention, given in the following subsections.

\subsubsection{Complex Permutations}
\label{complex-permutations}

An important and natural extension of the research within set-to-sequence is to apply it to challenges where the output sequence allows for repetition and exclusion of input set elements, thus going beyond traditional permutation learning \shortcite{Diallo2020}. As such, these \textit{complex permutations} present an additional challenge of dynamically predicting the optimal sequence length without sacrificing the second of the two aforementioned prerequisites for ML methods on sets, namely the requirement that the same model must be able to process finite input sets of any cardinality. It is not immediately clear how to achieve both properties, with multiple promising approaches gaining prominence, ranging from the addition of a confidence loss for long time-series prediction \cite{Harmon2019}, to the aforementioned adaptive computation time \shortcite{wu2020don}. For an overview of dynamic neural network methods in general, see \shortciteauthor{han2021dynamic} \citeyear{han2021dynamic}.

Important tasks involving a complex target permutation include predicting a configuration representing the assembly of off-the-shelf components into a functional system \cite{CarlsonSkalak1998UsingAE} and the composition of product offers into rendered catalogues \cite{Liao2004}. Additionally, an interesting area for further research would be to extend these complex permutation sequences to grids and lattices, as suggested by \citeauthor{Zhang2019permcomp} \citeyear{Zhang2019permcomp}, or even to graphs, expanding on the work of \shortciteauthor{serviansky2020set2graph} \citeyear{serviansky2020set2graph}.

\subsubsection{Differentiable Loss Functions}
\label{differentiable-loss-functions}

Another important area for further work is centered around the problem of framing the set-to-sequence challenges in such a way as to enable the use of differentiable loss functions. A naive but often practically effective approach is to frame the problem as categorization and use a cross-entropy loss, as discussed in \citeauthor{engilberge2019sodeep} \citeyear{engilberge2019sodeep}. However, it precludes meaningful distinction between pointing to an incorrect element that is very similar to the correct one and pointing to an entirely different incorrect element. 

Alternatively, if a ranking framework is applied, where a score is generated for each element and subsequently used to sort all elements into a new permutation, we gain access to well documented listwise losses, such as the ones successfully employed in the ListNet or \textit{ListMLE} \cite{Kumar2020} frameworks. Many metrics that would lend themselves to ordering challenges do not have defined derivatives for their entire domain. The development and testing of their smooth approximations is of great potential value, as seen in the works of \citeauthor{rolinek2020optimizing} \citeyear{rolinek2020optimizing} and \citeauthor{Blondel2020FastDS} \citeyear{Blondel2020FastDS}.

\section{Conclusion} \label{conclusion}

Set-to-sequence is currently established as a family of methods with an exceptionally wide range of applications. At its essence, it is a combination of three areas seeing a lot of attention within the machine learning and deep learning research communities, namely set-input problems, combinatorial optimization and structured prediction. As such, a number of methods that can further our understanding of this field originates from other areas of interest, sometimes without seeing immediate application to set-to-sequence challenges. Progress owes to advances in model architectures, parallel computing, hyperparameter optimization as well as the overall growing interest in applying machine learning solutions to a wider and wider gamut of industrial challenges.

We have presented an overview of set-to-sequence methods from the fields of machine learning and deep learning. The initial sections of this paper introduced the reader to the key concepts in relation to machine learning on sets, most notably the properties of permutation invariance, permutation equivariance and the requirement of handling inputs of varying lengths. In relation to this, in section \ref{set-encoding-methods}, the reader is introduced to a number of selected set-encoding model architectures, including a significantly detailed look at their underlying mathematical transformations. To facilitate comprehensive understanding, we compared these and other related methods through several summary tables presented in section \ref{method-comparison-set-encoding}.

Additionally, a survey of potential ordering methods has been provided in section \ref{orderingmethods}. This included the three primary ways of formalizing the output permutation in the set-to-sequence setting, namely listwise ranking, relaxations of permutation matrices and attention-based pointing. Once a permutation invariant set representation is obtained through one of the aforementioned set-encoding models, the ordering methods' calculations can be conditioned on this embedding to output a sequence of elements from the original input set, in one fully trainable deep learning framework. 

\section*{Acknowledgements}

This work was partly supported by an Innovation Fund Denmark research grant (number 9065-00017B) and by Tjek A/S. The authors would like to acknowledge Rasmus Pagh's assistance in the conceptualization of different set-to-sequence settings and general feedback on the early drafts of this paper.

\vskip 0.2in
\bibliography{Set-to-Sequence_Methods_in_Machine_Learning_-_a_Review}
\bibliographystyle{theapa}

\end{document}